\documentclass[onefignum,onetabnum]{siamonline190516}

\usepackage{lipsum}
\usepackage{amsfonts}
\usepackage{graphicx}
\usepackage{epstopdf}
\usepackage{algorithmic}
\ifpdf
  \DeclareGraphicsExtensions{.eps,.pdf,.png,.jpg}
\else
  \DeclareGraphicsExtensions{.eps}
\fi

\usepackage{enumitem}
\setlist[enumerate]{leftmargin=.5in}
\setlist[itemize]{leftmargin=.5in}

\newsiamremark{remark}{Remark}
\newsiamremark{hypothesis}{Hypothesis}
\crefname{hypothesis}{Hypothesis}{Hypotheses}
\newsiamthm{claim}{Claim}

\headers{Discriminant Dynamic Mode Decomposition}{N. Takeishi, K. Fujii, K. Takeuchi, and Y. Kawahara}

\title{Discriminant Dynamic Mode Decomposition\\for Labeled Spatio-Temporal Data Collections\thanks{Preprint.
\funding{The major part of this work was done when the first author was working at RIKEN Center for Advanced Intelligence Project. This work was supported by JSPS KAKENHI Grant Numbers JP19K21550, JP20H04075, JP19H04941, and JP18H03287; JST PRESTO Grant Number JPMJPR20C5; JST CREST Grant Number JPMJCR1913; and AMED Grant Number JP19dm0307009.}}}

\author{Naoya Takeishi\thanks{RIKEN Center for Advanced Intelligence Project, Tokyo, Japan
  (\email{naoya.takeishi@riken.jp}) and University of Applied Sciences and Arts Western Switzerland, Geneva, Switzerland.}
\and Keisuke Fujii\thanks{Graduate School of Informatics, Nagoya University, Aichi, Japan
  (\email{fujii@i.nagoya-u.ac.jp}) and RIKEN Center for Advanced Intelligence Project, Tokyo, Japan.}
\and Koh Takeuchi\thanks{Graduate School of Informatics, Kyoto University, Kyoto, Japan
  (\email{takeuchi@i.kyoto-u.ac.jp}) and RIKEN Center for Advanced Intelligence Project, Tokyo, Japan.}
 \and Yoshinobu Kawahara\thanks{Institute of Mathematics for Industry, Kyushu University, Fukuoka, Japan
  (\email{kawahara@imi.kyushu-u.ac.jp}) and RIKEN Center for Advanced Intelligence Project, Tokyo, Japan.}
}

\usepackage{amsopn}

\ifpdf
\hypersetup{
  pdftitle={Discriminant Dynamic Mode Decomposition},
  pdfauthor={N. Takeishi, K. Fujii, K. Takeuchi, and Y. Kawahara}
}
\fi

\usepackage{natbib}
\setcitestyle{numbers,open={[},close={]}}
\usepackage{url}
\usepackage{amssymb,amsmath,bm}
\usepackage{color}
\usepackage{subfig}
\usepackage{pifont}
\usepackage{enumitem}

\usepackage{pgfplots}
\pgfplotsset{compat=newest}
\usepackage{tikz}
\usetikzlibrary{backgrounds}
\usetikzlibrary{arrows}
\usetikzlibrary{arrows.meta}
\usetikzlibrary{shapes,shapes.geometric,shapes.misc}
\usetikzlibrary{external}
\pgfdeclarelayer{edgelayer}
\pgfdeclarelayer{nodelayer}
\pgfsetlayers{background,edgelayer,nodelayer,main}

\pgfplotsset{
    /pgfplots/xlabel near ticks/.style={
        /pgfplots/every axis x label/.style={
        at={(ticklabel cs:0.5)},anchor=near ticklabel
        }
    },
    /pgfplots/ylabel near ticks/.style={
        /pgfplots/every axis y label/.style={
        at={(ticklabel cs:0.5)},rotate=90,anchor=near ticklabel}
    }
}

\pgfplotsset{every axis/.append style={
    label style={font=\scriptsize},
    tick label style={font=\tiny},
}}

\pgfplotsset{
    legend image code/.code={
        \draw[mark repeat=2,mark phase=2]
            plot coordinates {
            (0cm,0cm)
            (0.15cm,0cm)        
            (0.3cm,0cm)         
        };%
    }
}

\newcommand{\tr}{\mathsf{T}}
\newcommand{\ct}{\mathsf{H}}
\newcommand{\pinv}{\dagger}
\newcommand{\trace}{\operatorname{tr}}

\DeclareMathOperator*{\maximize}{maximize}

\begin{document}

\maketitle

\begin{abstract}
    Extracting coherent patterns is one of the standard approaches towards understanding spatio-temporal data. Dynamic mode decomposition (DMD) is a powerful tool for extracting coherent patterns, but the original DMD and most of its variants do not consider label information, which is often available as side information of spatio-temporal data. In this work, we propose a new method for extracting distinctive coherent patterns from labeled spatio-temporal data collections, such that they contribute to major differences in a labeled set of dynamics. We achieve such pattern extraction by incorporating discriminant analysis into DMD. To this end, we define a kernel function on subspaces spanned by sets of dynamic modes and develop an objective to take both reconstruction goodness as DMD and class-separation goodness as discriminant analysis into account. We illustrate our method using a synthetic dataset and several real-world datasets. The proposed method can be a useful tool for exploratory data analysis for understanding spatio-temporal data.
\end{abstract}




\section{Introduction}
\label{intro}

Spatio-temporal data are ubiquitous in modern science and engineering, but they are often so complicated and high-dimensional that we can hardly read useful information from them directly.
One of popular approaches to understanding spatio-temporal data is to extract from them \emph{coherent patterns} across space and time with which we can summarize the characteristics of spatio-temporal data in some sense.
Such summarization can be useful not only for understanding underlying dynamics but also for crafting machine learning features and models, reduced order modeling, and designing controllers.

Extraction of spatio-temporal coherent patterns can be performed in different ways.
Many popular approaches are formulated as dimensionality reduction techniques including principal component analysis (PCA) and other matrix/tensor factorization techniques \citep[see, e.g.,][]{Jolliffe02,koren2009matrix,cichocki2009nonnegative}.
The use of PCA for coherent patterns extraction is indeed prevalent in a wide range of applications, and there have been countless numbers of use cases.
For example, we may characterize complex neural activities using a small number of spatio-temporal patterns computed by PCA \citep{Churchland12}.
Moreover, a method called proper orthogonal decomposition (POD), which is equivalent to PCA in principle, has been a standard technique for reduced order modeling in fluid mechanics.
Dynamic mode decomposition (DMD) is also known as a powerful tool to extract spatio-temporal coherent patterns \citep[see, e.g.,][and references therein]{DMDBook}.
DMD was proposed originally in the field of fluid mechanics \citep{Rowley09,Schmid10} and has been successfully utilized in a broad range of applications such as neuroscience \citep{Brunton16b}, epidemiology \citep{Proctor15}, power system analysis \citep{Susuki14}, and building maintenance design \citep{Georgescu12}.
DMD is favored mainly because it considers the dynamical properties of spatio-temporal data, which are usually ignored by PCA and POD.

%

In this work, we tackle the problem of extracting coherent patterns from \emph{labeled} spatio-temporal data collections.
In many practices of spatio-temporal data analysis, label information is available as side information.
For example, biological recordings, such as neural signals, are often accompanied with additional information of targets (e.g., types of applied stimuli).
Data from social phenomena, such as recordings of traffic and people, can often be labeled with information of calendar (e.g., days of week and holidays).
Such label information can be a useful clue for extracting meaningful coherent patterns from spatio-temporal data.

Data summarization considering label information has been addressed also in several other contexts.
For example, Fisher's linear discriminant \citep{Fisher36} is a classical and well-known method for dimensionality reduction of labeled data.
Supervised PCA \citep{Barshan11} is a variant of PCA that takes labels of each data-point into consideration.
In brain-computer interface studies, extraction of common spatial patterns \citep{Ramoser00} from a set of labeled neural signals is a popular tool for classification.
However, most existing methods for labeled data summarization ignore dynamical aspects of data, and thus these methods are not necessarily suitable for spatio-temporal data.

\begin{figure*}[t]
    \centering
    \begin{tikzpicture}
        \tikzstyle{background}=[fill=white, draw=black, shape=rectangle, align=center, minimum height=1.2cm]
        \tikzstyle{proposed}=[fill=white, draw=black, double distance=2pt, outer sep=2pt, shape=rectangle, align=center, minimum height=2cm]
        \tikzstyle{line}=[-]
        \tikzstyle{arrow}=[-{Latex[length=3mm]}]
        \begin{pgfonlayer}{nodelayer}
            \node [style=background] (0) at (-4.6, 0) {DMD\\(\cref{back:dynamics})};
            \node [style=background] (1) at (0, 0) {Grassmann kernels\\(\cref{back:grassmann})};
            \node [style=background] (2) at (4.6, 0) {Kernel disc. analysis\\(\cref{back:kfd})};
            \node [style=proposed] (3) at (4.6, 2.8) {\textbf{Discriminant DMD}\\(\cref{method:main} \\ \& \cref{method:example})};
            \node [style=proposed,minimum width=3.5cm] (4) at (-2.3, 2.8) {Dynamic mode\\subspace kernel\\(\cref{method:dmsk})};
            \node [] (5) at (-4.6, 1.2) {};
            \node [] (6) at (-2.3, 1.2) {};
            \node [] (7) at (0, 1.2) {};
        \end{pgfonlayer}
        \begin{pgfonlayer}{edgelayer}
            \draw [style=arrow] (4) to (3);
            \draw [style=arrow] (2) to (3);
            \draw [style=line] (0) to (5.center);
            \draw [style=line] (5.center) to (6.center);
            \draw [style=arrow] (6.center) to (4);
            \draw [style=line] (1) to (7.center);
            \draw [style=line] (7.center) to (6.center);
        \end{pgfonlayer}
    \end{tikzpicture}
    \caption{Construction of the paper. In \cref{back}, we introduce three technical building blocks: DMD, kernel discriminant analysis, and Grassmann kernels. We then define a kernel function on sets of dynamic modes in \cref{method:dmsk}. Finally, we formulate our proposal of discriminant DMD as a combination of the DMD-based kernel and kernel discriminant analysis in \cref{method:main}. We also show its numerical example in \cref{method:example}.}
    \label{fig:overview}
\end{figure*}
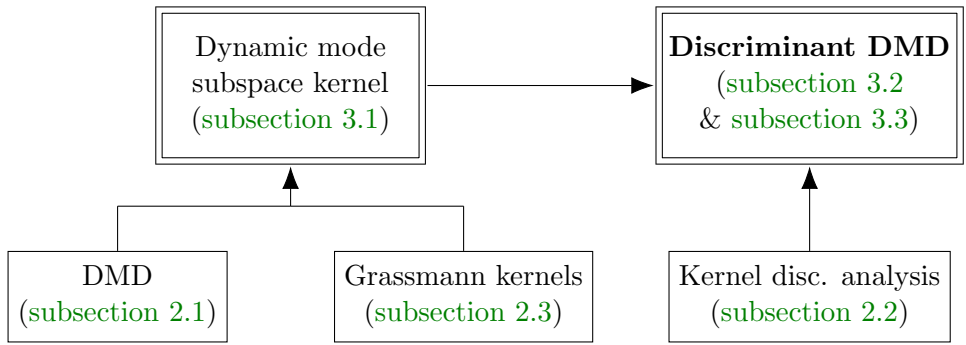

In this paper, we present a method to extract coherent patterns from labeled spatio-temporal data such that they reflect \emph{both} of the labels and the dynamics.
We build such an extraction method upon DMD.
Since the original DMD algorithm and the most existing variants cannot consider label information, we propose to incorporate discriminant analysis into DMD's formulation. 
Specifically, we combine an optimization formulation of DMD \citep{Chen12} with a parameter optimization method for kernel Fisher discriminant analysis \citep{You11}.
To this end, we need a kernel function to measure similarity between sets of dynamic modes extracted from different time-series episodes.
We design such a kernel function for sets of dynamics modes based on the notion of Grassmann kernels on subspaces \citep{Hamm08}.
Finally, we formulate an optimization problem to compute \emph{discriminant DMD}, with which we can obtain distinctive coherent patterns that contribute to major differences in spatio-temporal behaviors of different labels.

The remainder of the paper is structured as follows.
In \cref{back}, we review three techniques that constitute building blocks.
In \cref{method}, we then present the discriminant DMD with an illustrative numerical example.
\cref{fig:overview} summarizes the construction of \cref{back} and \cref{method}.
Afterward, we discuss the related work in \cref{related}.
We showcase some applied use cases of the proposed method with real-world datasets in \cref{expt}.
Finally, we give discussion in \cref{discussion} and conclude the paper in \cref{concl}.


\section{Background}
\label{back}

We review three lines of studies that constitute the technical building blocks of the proposed method, which combines DMD and discriminant analysis.
We first review these two notions in \cref{back:dynamics} and \cref{back:kfd}, respectively.
Afterward, we review the concept of Grassmann kernels in \cref{back:grassmann}, which plays a key role in combining the first two notions.


\subsection{Modal Decomposition of Dynamical Systems}
\label{back:dynamics}

The operator-theoretic perspective for analyzing dynamical systems is the first building block toward the purpose of this work.
We first introduce some basic notions of dynamical systems and their operator-theoretic view \citep{Mezic05} and then review the modal decomposition of dynamical systems by DMD \citep{Rowley09,Schmid10}.


\subsubsection{Dynamical Systems and Koopman Operators}

Variety of physical phenomena can be modeled as \emph{dynamical systems}.
In this paper, we primarily consider discrete-time dynamical systems like
\begin{equation}\label{eq:dynamical_system}
    v_{t+1} = f (v_t),
\end{equation}
where $v_t \in \mathcal{M}$ is a state vector, $\mathcal{M}$ is a state space, and $t \in \{0\}\cup\mathbb{N}$ is a time index.
When map $f$ is nonlinear, which is often the case, it is highly challenging to analyze the dynamical system's behavior directly in state space $\mathcal{M}$.

The operator-theoretic perspectives, such as ones based on the \emph{Koopman operator} \citep[see, e.g.,][]{Mezic05,Budisic12}, are attracting attention as an alternative to analyze nonlinear dynamical systems.
Consider a function on the state space, namely $g \colon \mathcal{M} \to \mathbb{R}$, which is referred to as an \emph{observable}.
We suppose $g$ is in some Hilbert space $\mathcal{G}$.
For the dynamical system in \cref{eq:dynamical_system}, the corresponding Koopman operator $\mathcal{K}$ is defined as
\begin{equation}\label{eq:koopman}
    \mathcal{K} g(v) = g ( f (v) ).
\end{equation}
Because $\mathcal{G}$ is a vector space, $\mathcal{K}$ is a \emph{linear} operator and infinite-dimensional in general.

An advantage of considering the Koopman operator appears as its spectral decomposition.
For simplicity of discussion, let us assume that the spectral decomposition of $\mathcal{K}$ is well-defined and that $\mathcal{K}$ has only simple point spectra.
Let $\lambda_j \in \mathbb{C}$ and $\varphi_j \colon \mathcal{M} \to \mathbb{C}$ be an eigenvalue and an eigenfunction of $\mathcal{K}$, respectively, i.e.,
\begin{equation}\label{eq:eigen}
    \mathcal{K} \varphi_j(v) = \lambda_j \varphi_j(v),
    \quad
    j \in \mathbb{N}.
\end{equation}
If $g$ is in the span of the (possibly countably many) eigenfunctions of $\mathcal{K}$, we have a decomposition of the observable:
\begin{equation}\label{eq:kmd1}
    \mathcal{K} g(v) = \sum_{j=1}^\infty \lambda_j \varphi_j(v) \tilde{w}_j(g),
\end{equation}
where $\tilde{w}_j(g)\in\mathbb{C}$ is a coefficient of the orthogonal projection of $g$ onto the span of $\varphi_j$.
By applying $\mathcal{K}$ to the both sides of \cref{eq:kmd1} repeatedly (see \cref{eq:koopman} and \cref{eq:eigen}), we obtain
\begin{equation}\label{eq:kmd2}
    g (v_t) = \sum_{j=1}^\infty \lambda_j^t \big( \varphi_j(v_0) \tilde{w}_j(g) \big),
\end{equation}
where $v_0\in\mathcal{M}$ is a initial condition.
The expression in \cref{eq:kmd1} or \cref{eq:kmd2} is called the Koopman mode decomposition (KMD) \citep{Budisic12} in literature and have been utilized for discovering meaningful spatio-temporal patterns of dynamics \citep[see, e.g.,][and references therein]{Budisic12,DMDBook}.


\subsubsection{Dynamic Mode Decomposition}

Dynamic mode decomposition (DMD) \citep{Rowley09,Schmid10} is a method to compute modal decomposition of dynamics.
DMD and its variants have been utilized in a wide range of domains \citep[see, e.g.,][and references therein]{DMDBook} partly because of its mathematical and algorithmic simplicity.
It also has a connection to KMD under some conditions \citep{Budisic12,Williams15a,Arbabi17,Korda18}.
We briefly review the idea and techniques of DMD.

\paragraph{Eigendecomposition-Based Formulation}
We present an abstracted description of a working algorithm of DMD based on pseudoinverse and eigendecomposition \citep{Tu14}.
Suppose we have a time-series episode $X = ( \bm{x}_1, \dots, \bm{x}_\tau )$ obtained from some dynamics, where $\bm{x}_t\in\mathbb{R}^p$ denotes an observation (also called a snapshot) at time $t$.
We compute the eigendecomposition of a matrix $\bm{A}$ such that
\begin{equation}
    \bm{A} = \bm{X}^+ (\bm{X}^-)^\pinv,
\end{equation}
where $\bm{X}^+$ and $\bm{X}^-$ are time-lagged data matrices, i.e.,
\begin{equation}\begin{aligned}
    \bm{X}^- &= \begin{bmatrix} \bm{x}_1 & \cdots & \bm{x}_{\tau-1} \end{bmatrix} \in\mathbb{R}^{p \times (\tau-1)}, \\
    \bm{X}^+ &= \begin{bmatrix} \bm{x}_2 & \cdots & \bm{x}_\tau \end{bmatrix} \in\mathbb{R}^{p \times (\tau-1)},
\end{aligned}\end{equation}
and $\cdot^\dagger$ denotes the Moore--Penrose pseudo inverse.
When $p \gg 1$, we often perform the dimensionality reduction of data to $r$ dimensions ($r<p$) using the singular value decomposition.
We then compute the eigendecomposition of the dimension-reduced version of $\bm{A}$, and finally the eigenvectors are projected back to the original $p$-dimensional space.
See \citep{Tu14} for the detailed procedures.

DMD's modal decomposition of a time-series episode is formulated using the spectra of the $\bm{A}$ matrix.
Let $\hat{\bm{w}}_j \in\mathbb{C}^p$ and $\bm{z}_j \in\mathbb{C}^p$ be a pair of the right- and left-eigenvectors of $\bm{A}$, respectively, corresponding to $\bm{A}$'s non-zero eigenvalue $\lambda_j \in \mathbb{C}$, for $j=1,\dots,p'$, where $p'$ is the number of the non-zero eigenvalues.
Without loss of generality, we suppose that they are normalized so that $\hat{\bm{w}}_j^\ct\bm{z}_{j'}=\delta_{j,j'}$, where $\delta_{j,j'}=1$ if $j=j'$ and $\delta_{j,j'}=0$ otherwise.
If $\bm{X}^-$ and $\bm{X}^+$ are linearly consistent \citep{Tu14} and $\lambda_1,\dots,\lambda_{p'}$ are distinct, we have
\begin{equation}\label{eq:dmd}
    \bm{x}_t = \sum_{j=1}^{p'} \lambda_j^{t-1} (\bm{z}_j^\ct\bm{x}_1)\hat{\bm{w}}_j.
\end{equation}
In \cref{eq:dmd}, each snapshot is represented as a weighted sum of the vectors, $\hat{\bm{w}}_1,\dots,\hat{\bm{w}}_{p'}$, which are termed \emph{dynamic modes}.
Note that the weights of each dynamic mode are given by the corresponding (complex) eigenvalues, $\lambda_1,\dots\lambda_{p'}$.
Hence, the $j$-th dynamic mode, $\bm{w}_j$, can be regarded as a spatial coherent pattern that exhibits oscillation with an angular frequency $\angle\lambda_j$ and a decay/growth rate $\vert\lambda_j\vert$.

There is a formal resemblance between KMD \cref{eq:kmd2} and DMD \cref{eq:dmd}.
In fact, we can find further connections between the two decomposition methods under some conditions.
Let us see one of them.
Suppose that a snapshot $\bm{x}$ is generated by a concatenation of the values of multiple observables $g_1,\dots,g_p \in \mathcal{G}$, i.e.,
\begin{equation}\label{eq:observables}
    \bm{x}_t = \bm{g}(v_t) = \begin{bmatrix} g_1(v_t) & \cdots & g_p(v_t) \end{bmatrix}^\tr.
\end{equation}
With this interpretation, Tu \emph{et al.} \citep{Tu14} have shown that if $\varphi_j$ is in the span of $\{g_1,\dots,g_p\}$, and the data is sufficiently rich, then pointwise evaluations of the eigenfunction are given by $\varphi_j(v_t) = \bm{z}_j^\ct \bm{x}_t$.
Moreover, the connection between variants of DMD and KMD has been discussed \citep{Williams15a,Arbabi17,Korda18}.
In this sense, DMD may be roughly understood as a way to approximate spectral components of the Koopman operator, though care must be taken when one needs such an interpretation.
We note that DMD is still a useful tool for dimensionality reduction and pattern extraction even if the connection to the Koopman operator spectrum is hardly established in practice.

\paragraph{Opimitzation-Based Formulation}
Apart from the eigendecomposition-based formulation, DMD can also be defined as an exponential curve fitting to multivariate time-series.
This paradigm is often termed optimized DMD \citep{Chen12} as it is based on nonlinear optimization.
Let
\begin{equation*}
    \bm{W} = \begin{bmatrix} \bm{w}_1 & \cdots & \bm{w}_r \end{bmatrix}
    \in \mathbb{C}^{p \times r}
\end{equation*}
be a matrix comprising $r$ ($\leq p$) dynamic modes (we can think that $\bm{w}_j$ corresponds to $(\bm{z}_j^\ct\bm{x}_1)\hat{\bm{w}}_j$ in \cref{eq:dmd} in its role).
Let $\bm{V}_{\lambda_{1:r}} \in \mathbb{C}^{r \times \tau}$ denote a Vandermonde matrix parameterized by a set $\lambda_{1:r}=\{ \lambda_1, \dots, \lambda_j \}$ as
\begin{equation}\label{eq:vandermonde}
    \bm{V}_{\lambda_{1:r}} =
    \begin{bmatrix}
        1 & \lambda_1 & \lambda_1^2 & \cdots & \lambda_1^{\tau-1} \\
        1 & \lambda_2 & \lambda_2^2 & \cdots & \lambda_2^{\tau-1} \\
        \vdots & \vdots & \vdots & & \vdots \\
        1 & \lambda_r & \lambda_r^2 & \cdots & \lambda_r^{\tau-1}
    \end{bmatrix}
    \in \mathbb{C}^{r \times \tau}.
\end{equation}
Then, optimized DMD is formulated as an optimization problem
\begin{equation}\label{eq:optdmdprob}
    \underset{\bm{W},\lambda_{1:r}}{\text{minimize}} ~~ \Vert \bm{X} - \bm{W}\bm{V}_{\lambda_{1:r}} \Vert_F^2.
\end{equation}
The objective in \cref{eq:optdmdprob} measures the squared error of the reconstruction $\bm{W}\bm{V}_{\lambda_{1:r}}$ with the dynamic modes.
Conceptually it corresponds to the error between the left- and right-hand sides of \cref{eq:dmd} of the eigendecomposition-based formulation.

The optimization problem in \cref{eq:optdmdprob} can be addressed with the strategy called \emph{variable projection} \citep{Golub73,Askham18} because it is a linear least squares problem with regard to $\bm{W}$.
That is, if we fix $\lambda_{1:r}$ at some value $\lambda_{1:r}^*$, the corresponding optimal $\bm{W}^*$ is immediately determined by
\begin{equation}\label{eq:varpro}
    \bm{W}^* = \bm{X} \big( \bm{V}_{\lambda_{1:r}^*} \big)^\pinv.
\end{equation}
Hence, \Cref{eq:optdmdprob} reduces to another nonlinear problem:
\begin{equation}\label{eq:optdmd}
    \underset{\lambda_{1:r}}{\text{minimize}} ~~ \left\Vert \bm{X} - \bm{X} \bm{V}_{\lambda_{1:r}}^\pinv \bm{V}_{\lambda_{1:r}} \right\Vert_F^2.
\end{equation}
See \citep{Askham18} for details on this formulation.
Later in \cref{method}, we will show that formulating DMD as an optimization problem is useful in developing our proposal to utilize label information.


\subsection{Discriminant Analysis}
\label{back:kfd}

Discriminant analysis forms the second building block of our method, and here we review a basic form and extensions.
First, we briefly introduce the Fisher's linear discriminant \citep{Fisher36} and its nonlinear counterpart using the kernel trick \citep{Mika99}.
Then, we review a method \citep{You11} to optimize kernel parameters of the kernelized discriminant analysis.


\subsubsection{Fisher's linear discriminant}

Fisher's linear discriminant \citep{Fisher36} is a classical supervised dimensionality reduction method and finds a linear transform of features so that different classes are well separated.
It is defined as maximization of a ratio of between- and within-class variances.
Suppose we have a set of data points\footnote{We slightly violate the notations in \cref{back:dynamics}; here, $\bm{x}$ is just a data point in general and does not necessarily constitute a sequence as a snapshot.} with binary labels, $\{(\bm{x}_i, y_i) \mid i=1,\dots,n\}$, where $\bm{x}_i \in \mathbb{R}^p$ and $y_i \in \{1,2\}$.
Then, Fisher's linear discriminant can be formulated as an optimization problem
\begin{equation}\label{eq:lda}
    \maximize_{\bm{b} \in \mathbb{R}^p} ~~ \frac{\bm{b}^\tr \bm{S}_\mathrm{B} \bm{b}}{\bm{b}^\tr \bm{S}_\mathrm{W} \bm{b}}.
\end{equation}
In \cref{eq:lda}, $\bm{S}_\mathrm{B}$ and $\bm{S}_\mathrm{W}$ denote between- and within-class scatter matrices defined as
\begin{equation}\label{eq:scatters}\begin{aligned}
    \bm{S}_\mathrm{B} &= \frac1n \sum_{l=1,2} n_l (\bm\mu_l - \bm\mu_\text{all}) (\bm\mu_l - \bm\mu_\text{all})^\tr,
    \\
    \bm{S}_\mathrm{W} &= \frac1n \sum_{l=1,2} \sum_{i \mid y_i=l} (\bm{x}_i - \bm\mu_l) (\bm{x}_i - \bm\mu_l)^\tr,
\end{aligned}\end{equation}
where $n_l$ and $\bm\mu_l$ denote the number and the average of data points with label $y=l$ (for $l=1,2$), respectively, i.e.,
\begin{equation}\label{eq:means}
    n_l = \vert \{i | y_i=l, \, i=1,\dots,n\} \vert
    \quad\text{and}\quad
    \bm\mu_l = \frac1{n_l} \sum_{i \mid y_i=l} \bm{x}_i.
\end{equation}
$\bm\mu_\text{all}$ denotes the average of all the data points, $\bm\mu_\text{all} = \frac1n \sum_{i=1}^n \bm{x}_i$.
\Cref{eq:lda} can be solved as a generalized eigenvalue problem.


\subsubsection{Kernel Fisher Discriminant Analysis}

Mika \emph{et al.} \citep{Mika99} formulated a nonlinear extension of Fisher's linear discriminant using the kernel trick, namely \emph{kernel Fisher discriminant} (KFD).
Let $k \colon \mathbb{R}^p \times \mathbb{R}^p \to \mathbb{R}$ be a kernel function and $\phi \colon \mathbb{R}^p \to \mathcal{H}$ be the corresponding feature map, where $\mathcal{H}$ denotes the reproducing kernel Hilbert space (RKHS) corresponding to $k$.
Then, in a two-class case, KFD is formulated as the problem:
\begin{equation}
    \maximize_{\bm{b}^\phi \in \mathcal{H}} ~~ \frac{(\bm{b}^\phi)^\tr \bm{S}^\phi_\mathrm{B} \bm{b}^\phi}{(\bm{b}^\phi)^\tr \bm{S}^\phi_\mathrm{W} \bm{b}^\phi},
\end{equation}
where $\bm{S}_\mathrm{B}^\phi$ and $\bm{S}_\mathrm{W}^\phi$ denote between- and within-class scatter matrices in feature space $\mathcal{H}$, respectively.
These scatter matrices are defined analogously to \cref{eq:scatters} as follows:
\begin{equation}\label{eq:scatters_rkhs}\begin{aligned}
    \bm{S}^\phi_\mathrm{B} &= \frac1n \sum_{l=1,2} n_l (\bm\mu^\phi_l - \bm\mu^\phi_\text{all}) (\bm\mu^\phi_l - \bm\mu^\phi_\text{all})^\tr,
    \\
    \bm{S}^\phi_\mathrm{W} &= \frac1n \sum_{l=1,2} \sum_{i \mid y_i=l} (\phi(\bm{x}_i) - \bm\mu^\phi_l) (\phi(\bm{x}_i) - \bm\mu^\phi_l)^\tr,
\end{aligned}\end{equation}
where $\bm\mu^\phi_l$ denote the sample mean of the data points of the $l$-th class ($l\in[1,c]$) in $\mathcal{H}$, that is,
\begin{align}
    \bm\mu^\phi_l &= \frac1{n_l} \sum_{i \mid y_i=l} \phi(\bm{x}_i).
\end{align}
$\bm\mu^\phi_\text{all}$ denotes the average of all the data points in $\mathcal{H}$, $\bm\mu^\phi_\text{all} = \frac1n \sum_{i=1}^n \phi(\bm{x}_i)$.

We do not describe the solution of KFD because we do not need to solve it explicitly.
Instead, we use it for providing a criterion of the goodness of discriminant through a kernel parameter optimization scheme for KFD, which is reviewed below.


\subsubsection{Kernel Parameter Optimization for KFD}

It is often difficult to predetermine the parameters of a kernel used in KFD.
You \emph{et al.} \citep{You11} proposed a method for optimizing kernel parameters of KFD.
Instead of the two-class case, let us consider a $c$-class case in general.
The method of You \emph{et al.} \citep{You11} tries to optimize kernel parameters via the following problem:
\begin{equation}\label{eq:Q0}
    \maximize ~~ Q_1 Q_2,
\end{equation}
\begin{align}
    Q_1 &= \frac2{c(c-1)} \sum_{l=1}^{c-1} \sum_{l'=l+1}^c
    \frac{
        \trace(\bm\Sigma_l^\phi \bm\Sigma_{l'}^\phi)
    }{
        \trace(\bm\Sigma_l^\phi \bm\Sigma_l^\phi) + \trace(\bm\Sigma_{l'}^\phi \bm\Sigma_{l'}^\phi)
    }, \label{eq:Q1}
    \\
    Q_2 &= \sum_{l=1}^{c-1} \sum_{l'=l+1}^c \frac{n_l n_{l'}}{n^2} \Vert \bm\mu^\phi_l - \bm\mu^\phi_{l'} \Vert_, \label{eq:Q2}
\end{align}
where $\bm\Sigma^\phi_l$ denotes the sample covariance matrix of the data points of the $l$-th class in $\mathcal{H}$, that is,
\begin{equation}
    \bm\Sigma^\phi_l = \frac1{n_l} \sum_{i \mid y_i=l} (\phi(\bm{x}_i)-\bm\mu^\phi_l) (\phi(\bm{x}_i)-\bm\mu^\phi_l)^\tr.
\end{equation}
$Q_1$ measures the homoscedasticity of data points of each class in the feature space, and $Q_2$ measures the class-separability in the feature space.
By maximizing the product of these objectives, $Q_1 Q_2$, we can adjust the kernel parameters so that the data points in the kernel feature space tend to be linearly separable.


\subsection{Grassmann Kernels on Subspaces}
\label{back:grassmann}

The third building block of the proposed method is the notion of Grassmann kernels on subspaces \citep[see, e.g.,][]{Hamm08}.
Grassmann kernels enable us to compute similarity between sets of vectors even if we do not have correspondence between the vectors in different sets.
In this work, we specifically use it to define a similarity between two sets of dynamic modes.

One of the kernels introduced by Hamm and Lee \citep{Hamm08} is called the \emph{projection kernel}.
Let $\mathcal{B}$ denote some subspace of Euclidean space, and let $\mathcal{B}_1, \mathcal{B}_2 \subset \mathcal{B}$ be subspaces of $\mathcal{B}$.
The projection kernel between those two subspaces, $k_\text{P}(\mathcal{B}_1,\mathcal{B}_2)$, is defined as
\begin{equation*}
    k_\text{P}(\mathcal{B}_1,\mathcal{B}_2) = \Vert \bm{B}_1^\ct\bm{B}_2 \Vert_F^2,
\end{equation*}
where $\bm{B}_i$ is the matrix whose columns comprise an orthonormal basis of $\mathcal{B}_i$ for $i=1,2$.
In fact, the projection kernel induces a metric based on principal angles between subspaces, which is called the projection metric \citep{Hamm08}.
It is defined as $d_\text{P}(\mathcal{B}_1,\mathcal{B}_2) = (\sum_d \sin^2 \alpha_d)^{1/2}$, where $\alpha_d$ denotes the $d$-th principal angle between $\mathcal{B}_1$ and $\mathcal{B}_2$.
Hence, we can understand the projection kernel as a similarity measure based on principal angles between subspaces.



\section{Proposed Method}
\label{method}

\begin{figure}[p]
    \centering
    \includegraphics[clip,width=\linewidth]{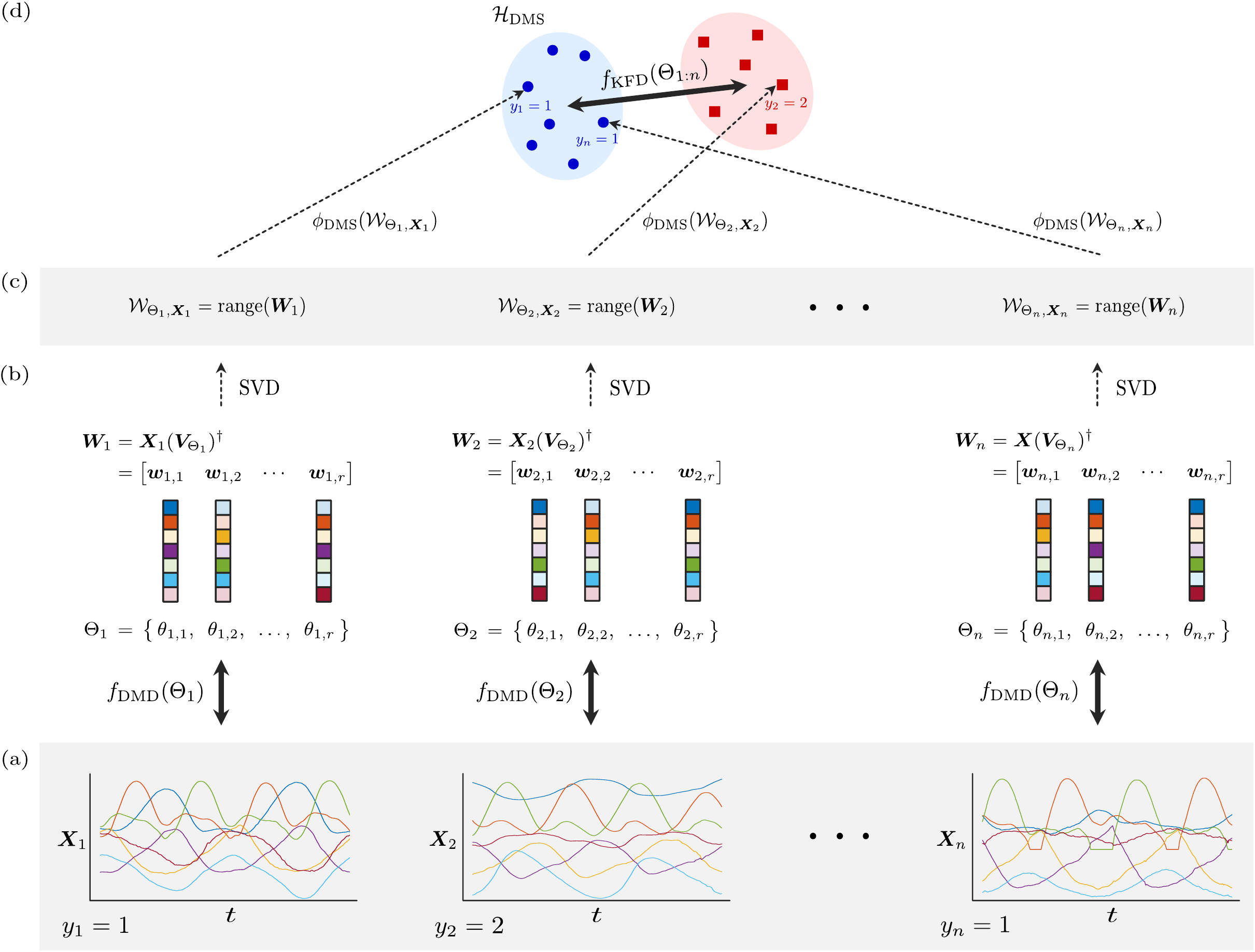}
    \caption{Schematic diagram of the two objectives, $f_\text{DMD}$ and $f_\text{KFD}$, which appear in the optimization of the proposed method (see \cref{eq:problem}). (a) We have pairs of time-series $\bm{X}_i$ and label $y_i$ (for $i=1,\dots,n$) as a dataset (see \cref{eq:dataset}--\cref{eq:label}). (b) Given such data, we aim to compute $\Theta_i$ and $\bm{W}_i$ for $i=1,\dots,n$. The elements of $\Theta$ and the columns of $\bm{W}$ correspond to DMD's eigenvalues and dynamic modes, respectively. Note that the to-be-optimized variable is only $\Theta$ because $\bm{W}$ is straightforwardly computed given $\Theta$ (see \cref{eq:varpro_again}). One of the objectives to optimize $\Theta$ is $f_\text{DMD}(\Theta_i)$, which measures the fitting error between $\bm{X}_i$ and $\bm{W}\bm{V}_{\Theta_i}$ (see \cref{eq:dmdloss}). (c) As correspondence between columns of different $\bm{W}$'s cannot be established in general, we compute the similarity of $\bm{W}$'s via their column space $\operatorname{range}(\bm{W})$, which can be computed by SVD of $\bm{W}$. We define a kernel function $k_\text{DMS}$ on such subspaces (see \cref{eq:kdms}). (d) The corresponding feature map is denoted as $\phi_\text{DMS}$. Another objective to optimize $\Theta$ is $f_\text{KFD}(\Theta_{1:n})$, which measures the homoscedasticity of data points in each class and their class-separability in an RKHS $\mathcal{H}_\text{DMS}$ induced by $k_\text{DMS}$ (see \cref{eq:kfdloss}). Note that the bold arrow at $f_\text{KFD}(\Theta_{1:n})$ in the figure only depicts the class-separability as the homoscedasticity is more difficult to depict.}
    \label{fig:schematics}
\end{figure}

In this section, we describe the details of our proposal of a method for discovering distinctive coherent patterns from labeled spatio-temporal data collections.
The main idea of the proposed method lies in combining DMD (reviewed in \cref{back:dynamics}) and KFD (\cref{back:kfd}) using a Grassmann kernel (\cref{back:grassmann}).
To this end, first we prepare a kernel to measure similarity between sets of dynamic modes in \cref{method:dmsk}.
Then, we formulate the main optimization problem \cref{eq:problem} to compute distinctive dynamic modes in \cref{method:main}.
Finally, we show numerical examples on synthetic data in \cref{method:example}.
The overall schematics of the proposed method is shown in \cref{fig:schematics}.


\subsection{Dynamic Mode Subspace Kernels for Time-Series Episodes}
\label{method:dmsk}

In computing dynamic modes, most variants of DMD cannot consider label information of time-series episodes.
We address this issue by developing a new method combining virtues of DMD and KFD.
We would like to compute dynamic modes that are good both in the sense of DMD (i.e., curve fitting \cref{eq:optdmd}) and in the sense of KFD (i.e., class homoscedasticity and separability \cref{eq:Q0}--\cref{eq:Q1}).
To this end, we should prepare a kernel function between \emph{sets of} dynamic modes.
An issue here is that in general, we cannot establish explicit correspondences between dynamic modes computed from two different datasets.
Hence, simply computing similarity between dynamic mode matrices $\bm{W}$ cannot be a valid approach.
Instead, we suggest using the projection kernel that computes similarity between subspaces spanned by sets of dynamic modes.

First, let us formalize a subspace spanned by dynamic modes, namely a \emph{dynamic mode subspace} (DMS):
\begin{definition}[Dynamic mode subspace]\label{def:dmsubspace}
    Let $\bm{X}\in\mathbb{C}^{p \times \tau}$ be a data matrix whose columns comprise $p$-dimensional snapshots of a time-series episode of length $\tau$.
    Let $\Theta=\theta_{1:r}=\{\theta_j \in \mathbb{C} \mid j=1,\dots,r\}$ be a set of $r$ distinct complex values ($0 < r \leq p$).
    We define the \emph{dynamic mode subspace} $\mathcal{W}_{\Theta,\bm{X}}$ of $\bm{X}$ corresponding to $\Theta=\theta_{1:r}$ as
    \begin{align}
        \mathcal{W}_{\Theta,\bm{X}} &= \operatorname{range} \big( \bm{W} \big) \subset \mathbb{C}^p,
        \\
        \bm{W} &= \bm{X} \big( \bm{V}_{\Theta} \big)^\pinv, \label{eq:varpro_again}
    \end{align}
    where $\bm{V}_{\Theta}$ is a Vandermonde matrix defined like in \cref{eq:vandermonde} with set $\Theta=\theta_{1:r}$.
\end{definition}

We note that, in \cref{def:dmsubspace}, the columns of $\bm{W} = \bm{X} ( \bm{V}_{\Theta} )^\pinv \in \mathbb{C}^{p \times r}$ are the dynamic modes computed from data $\bm{X}$ with $\Theta$ being the set of DMD eigenvalues (in the sense of optimized DMD \citep{Chen12}), using the variable projection technique in \cref{eq:varpro}.
In other words, dynamic modes appear in the definition of DMS only implicitly, and we can parametrize a DMS only with a set of $r$ complex scalars, $\Theta$, instead of a dynamic mode matrix $\bm{W}$ of size $p \times r$.
This property of DMS plays a key role in the optimization problem in \cref{method:main}.
We then define the projection kernel on DMS as follows.
\begin{definition}[Projection kernel on DMS]\label{def:kernel}
    Let $\bm{X}_1$ and $\bm{X}_2$ be two data matrices, and let $\Theta_1$ and $\Theta_2$ be two sets of $r$ distinct complex values.
    Let $\mathcal{W}_{\Theta_1,\bm{X}_1}$ and $\mathcal{W}_{\Theta_2,\bm{X}_2}$ be their DMS's corresponding to $\Theta_1$ and $\Theta_2$, respectively.
    The projection kernel \citep{Hamm08} between the two subspaces, $\mathcal{W}_{\Theta_1,\bm{X}_1}$ and $\mathcal{W}_{\Theta_2,\bm{X}_2}$, is
    \begin{equation}\label{eq:kdms}\begin{aligned}
        k_{\text{DMS}}(\mathcal{W}_{\Theta_1,\bm{X}_1}, \mathcal{W}_{\Theta_2,\bm{X}_2})
        &= \Vert \bm{B}_{\Theta_1,\bm{X}_1}^\ct\bm{B}_{\Theta_2,\bm{X}_2} \Vert_F^2,
    \end{aligned}\end{equation}
    where the columns of $\bm{B}_{\Theta_1,\bm{X}_1}$ and $\bm{B}_{\Theta_2,\bm{X}_2}$ comprise an orthonormal basis of $\mathcal{W}_{\Theta_1,\bm{X}_1}$ and $\mathcal{W}_{\Theta_1,\bm{X}_1}$, respectively.
\end{definition}
We denote the feature map corresponding to kernel function $k_\text{DMS}$ by
\begin{equation*}
\phi_\text{DMS}: \mathcal{W}_{\Theta,\bm{X}} \mapsto \phi_\text{DMS}(\mathcal{W}_{\Theta,\bm{X}}).
\end{equation*}
Moreover, we denote the RKHS induced by $k_\text{DMS}$ by $\mathcal{H}_\text{DMS}$.
We can compute $\bm{B}_{\Theta,\bm{X}}$ by the singular value decomposition (SVD) of $\bm{W}=\bm{X}(\bm{V}_\Theta)^\pinv$; that is, $\bm{B}_{\Theta,\bm{X}}$ can be a matrix comprising left singular vectors of $\bm{X}(\bm{V}_\Theta)^\pinv$.

The derivative of $k_\text{DMS}$ with regard to an element of $\Theta_1$ or $\Theta_2$ can be computed as follows.
Let $\theta_{1,j}$ denote the $j$-th element of $\Theta_1$.
Then\footnote{When $\bm{Y}$ is an $n \times m$ matrix, we define both $\partial x / \partial \bm{Y}$ and $\partial \bm{Y} / \partial x$ as an $n \times m$ matrix.},
\begin{equation}
    \frac{\partial k_{\text{DMS}}(\mathcal{W}_{\Theta_1,\bm{X}_1}, \mathcal{W}_{\Theta_2,\bm{X}_2})}{\partial \theta_{1,j}}
    = \operatorname{sum} \left(
        \frac{\partial k_{\text{DMS}}}{\partial \bm{V}_{\Theta_1}} \circ
        \frac{\partial \bm{V}_{\Theta_1}}{\partial \theta_{1,j}}
    \right),
\end{equation}
where $\circ$ is the Hadamard product, and $\operatorname{sum}(\bm{Y})$ means the summation over all the elements of matrix $\bm{Y}$.
Suppose that $\bm{B}_{\Theta_1,\bm{X}_1}$ is obtained by the SVD of $\bm{X}_1 (\bm{V}_{\Theta_1})^\pinv$.
Then, we have
\begin{multline}
    \left( \frac{\partial k_{\text{DMS}}}{\partial \bm{V}_{\Theta_1}} \right)^\tr
    =
    (\bm{V}_{\Theta_1})^\pinv (\bm{V}_{\Theta_1}^\ct)^\pinv \bm{C}_1^\ct (\bm{I} - \bm{V}_{\Theta_1} (\bm{V}_{\Theta_1})^\pinv)^\ct \\
    + (\bm{I} - (\bm{V}_{\Theta_1})^\pinv \bm{V}_{\Theta_1})^\ct \bm{C}_1^\ct (\bm{V}_{\Theta_1}^\ct)^\pinv (\bm{V}_{\Theta_1})^\pinv
    - (\bm{V}_{\Theta_1})^\pinv \bm{C}_1 (\bm{V}_{\Theta_1})^\pinv,
\end{multline}
where
\begin{equation}
    \bm{C}_1 = (\bm{X}_1 (\bm{V}_{\Theta_1})^\pinv)^\pinv \bm{B}_{\Theta_2,\bm{X}_2} \bm{B}_{\Theta_2,\bm{X}_2}^\ct (\bm{I} - \bm{B}_{\Theta_1,\bm{X}_1} \bm{B}_{\Theta_1,\bm{X}_1}^\ct) \bm{X}_1.
\end{equation}
The above discussion analogously applies to the computation of $\partial k_\text{DMS}/\partial \theta_{2,j}$.




\subsection{Discriminant Dynamic Mode Decomposition}
\label{method:main}

This section describes our problem setting more formally and the main part of the proposed method.


\subsubsection{Problem Setting}

We suppose that we have a \emph{dataset} $\mathcal{D}$:
\begin{equation}\label{eq:dataset}
    \mathcal{D} = \{ (X_1, y_1), \dots, (X_n,y_n) \},
\end{equation}
whose element is a pair of an \emph{episode} $X_i$ and a \emph{label} $y_i$.
An episode $X_i$ is a sequence of observation vectors, i.e.,
\begin{equation}
    X_i = ( \bm{x}_{i,1}, \dots, \bm{x}_{i,\tau_i} ),
\end{equation}
where $\bm{x}_{i,t} \in \mathbb{R}^p$ ($t=1,\dots,\tau_i$) is an observation at time-step $t$ and also called a \emph{snapshot}.
$\tau_i$ is the length of the episode $X_i$, and $p$ is the dimensionality of a snapshot.
We may denote the episode $X_i$ also by a matrix
\begin{equation}
    \bm{X}_i = \begin{bmatrix}
    \bm{x}_{i,1} & \cdots & \bm{x}_{i,\tau_i}
    \end{bmatrix} \in \mathbb{R}^{p \times \tau_i}.
\end{equation}
In a dataset, each episode is associated with a label
\begin{equation}\label{eq:label}
    y_i \in \{1, \ldots, c \}.
\end{equation}
The label of an episode represents some distinctive property (within $c$ classes) of the episode.

Given such a collection of labeled time-series episodes as a dataset, we would like to obtain sets of complex values $\Theta_i=\{ \theta_{i,1},\dots,\theta_{i,r} \}$ (for $i=1,\dots,n$) such that:
\begin{enumerate}
    \item The labeled episodes are well separated in KFD's sense; and
    \item The snapshots of each episode are well fitted in the sense of the optimized DMD's sense.
\end{enumerate}
Here we note that we only deal with the sets of time-evolution parameters $\Theta_i$ (for $i=1,\dots,n$) as optimization variables because once they are determined, we can immediately compute the corresponding dynamic modes $\bm{W}$ by substituting $\theta_{i,1:r}$ to $\lambda^*_{1:r}$ in \cref{eq:varpro}.


\subsubsection{Formulation}

We formulate the proposed method, \emph{discriminant DMD}, as the following optimization problem:
\begin{equation}\label{eq:problem}\begin{gathered}
    \underset{\Theta_{1:n}}{\text{minimize}} ~~
    \frac{n^{-1} \sum_{i=1}^n f_\text{DMD}(\Theta_i)}{f_\text{KFD}(\Theta_{1:n})^\alpha + \epsilon},
\end{gathered}\end{equation}
where $\alpha \geq 0$ is a hyperparameter that balances the importance of $f_\text{KFD}$ and $f_\text{DMD}$, and $\epsilon \geq 0$ is a small number for numerical stability.
The two terms in \cref{eq:problem} are to achieve the purposes of the discriminant DMD listed above and defined as follows.

\paragraph{KFD term}
For letting the DMS projection kernel well separate data points (i.e., labeled episodes), we try to maximize
\begin{equation}\label{eq:kfdloss}
    f_\text{KFD}(\Theta_{1:n}) = Q_1 Q_2,
\end{equation}
with respect to $\Theta_{1:n}$.
We define $Q_1$ and $Q_2$ like \cref{eq:Q1} and \cref{eq:Q2}, respectively, with $\phi$ being $\phi_\text{DMS}$, the feature map corresponding to $k_{\text{DMS}}$.
Hence, $Q_1$ and $Q_2$ depend on $\Theta_{1:n}$ via the arguments of $k_\text{DMS}$.
Again let us emphasize that a data point of KFD here corresponds to an episode (\emph{not} a snapshot), so kernel function $k_\text{DMS}$ will be called $O(n^2)$ times.

\paragraph{DMD term}
For making the elements of $\Theta_i$ good for fitting $\bm{X}_i$ in the optimized DMD's sense (i.e., \cref{eq:optdmd}), for $i=1,\dots,n$, we try to minimize
\begin{equation}\label{eq:dmdloss}
    f_\text{DMD}(\Theta_i) = \frac1{\tau_i} \big\Vert \bm{X}_i - \bm{X}_i (\bm{V}_{\Theta_i})^\pinv \bm{V}_{\Theta_i} \big\Vert_F^2.
\end{equation}
It is exactly the same with the one used in the optimized DMD with the variable projection \citep{Askham18}.

\begin{algorithm}[t]
    \renewcommand{\algorithmicrequire}{\textbf{Input:}}
    \renewcommand{\algorithmicensure}{\textbf{Output:}}
    \def\NoNumber#1{{\def\alglinenumber##1{}\State #1}\addtocounter{ALG@line}{-1}}
    \caption{Discriminant DMD}
    \label{alg:main}
    \begin{algorithmic}[1]
        \REQUIRE Dataset $\mathcal{D} = \{ (X_i, y_i) \mid i=1,\dots,n \}$ (i.e., a set of pairs of an episode $X_i = ( \bm{x}_{i,1}, \dots, \bm{x}_{i,\tau_i} )$ and a label $y_i$) and hyperparameter $\alpha$
        \ENSURE Sets of time-evolution parameters $\{\Theta_i \mid i=1,\dots,n\}$ (i.e., what corresponds to DMD's eigenvalues) and sets of distinctive dynamic modes $\{ \bm{W}_i \mid i=1,\dots,n \}$ (each column of $\bm{W}_i$ is a dynamic mode)
        \FOR{$i = 1 \, \ldots \, n$}
            \STATE $\Theta_i, \bm{W}_i \leftarrow \text{StandardDMD}(X_i)$ \algorithmiccomment{\citep{Tu14} or \citep{Chen12,Abraham17}; see \cref{back:dynamics}}
        \ENDFOR
        \STATE $\Theta_1,\dots,\Theta_n \leftarrow \arg\min_{\Theta_1:n} \frac{n^{-1} \sum_{i=1}^n f_\text{DMD}(\Theta_i)}{f_\text{KFD}(\Theta_{1:n})^\alpha + \epsilon}$ \algorithmiccomment{See \cref{eq:problem}, \cref{eq:kfdloss}, and \cref{eq:dmdloss}}
        \FOR{$i = 1 \, \ldots \, n$}
            \STATE $\bm{W}_i \leftarrow$ compute \cref{eq:varpro} with $\lambda^*$ being $\Theta_i$
        \ENDFOR
    \end{algorithmic}
\end{algorithm}



\subsubsection{Optimization}

Local solutions of \cref{eq:problem} can be found using gradient methods.
In the numerical experiments below, we used a quasi-Newton optimizer with initial values computed by the standard DMD algorithm based on eigendecomposition \citep{Tu14}.
We present the details on the gradient of the objective in \cref{appendix:grad}.
The overall algorithm of the discriminant DMD is summarized in \cref{alg:main}.


\subsection{Numerical Example}
\label{method:example}

\begin{figure*}[t]
    \centering
    \subfloat[]{\begin{minipage}[t]{68mm}\vspace*{0pt}\centering
        {\scriptsize\begin{tabular}{ccc}
            \includegraphics[clip,width=1.8cm]{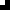} &
            \includegraphics[clip,width=1.8cm]{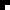} &
            \includegraphics[clip,width=1.8cm]{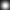} \\
            $\bm{w}_{\mathrm{d},y=1}$ & $\bm{w}_{\mathrm{d},y=2}$ & $\bm{w}_\mathrm{c}$
        \end{tabular}}
        \label{fig:example:data}
    \end{minipage}}
    \hspace{10mm}
    \subfloat[]{\begin{minipage}[t]{46mm}\vspace*{0pt}\centering
        \begin{tikzpicture}[inner sep=2pt, outer sep=0pt]
            \pgfplotsset{width=5cm,height=3cm}
            \begin{axis}[
                xticklabel style={
                    /pgf/number format/fixed,
                    /pgf/number format/precision=2
                },
                scaled x ticks=false,
                ylabel near ticks,
                xlabel near ticks,
                xlabel={$1/f_\text{KFD}$},
                ylabel={$f_\text{DMD}$},
                label style={font=\scriptsize}
            ]
                \addplot table [only marks, x=kfd, y=dmd] {expt_image/disc_loss_recip.txt};
            \end{axis}
        \end{tikzpicture}
        \label{fig:example:loss}
    \end{minipage}}
    \par
    \vspace*{2mm}
    \subfloat[]{
        \begin{minipage}[t]{38mm}\vspace*{0pt}\centering
            \includegraphics[clip,width=15mm]{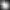}
            \hspace{0.5mm}
            \includegraphics[clip,width=15mm]{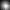}
            \par\vspace*{4mm}
            \begin{tikzpicture}[inner sep=0pt, outer sep=0pt]
                \pgfplotsset{width=48mm}
                \begin{axis}[xticklabels={,,},yticklabels={,,}]
                    \addplot table [only marks] {expt_image/exact_cmds_1.txt};
                    \addplot table [only marks] {expt_image/exact_cmds_2.txt};
                \end{axis}
            \end{tikzpicture}
            {\scriptsize standard DMD\\[-4pt]\citep[e.g.,][]{Rowley09}}
        \end{minipage}
        \hspace{3mm}
        \begin{minipage}[t]{39mm}\vspace*{0pt}\centering
            \includegraphics[clip,width=15mm]{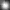}
            \hspace{0.5mm}
            \includegraphics[clip,width=15mm]{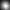}
            \par\vspace*{4mm}
            \begin{tikzpicture}[inner sep=0pt, outer sep=0pt]
                \pgfplotsset{width=48mm}
                \begin{axis}[xticklabels={,,},yticklabels={,,}]
                    \addplot table [only marks] {expt_image/sup_cmds_1.txt};
                    \addplot table [only marks] {expt_image/sup_cmds_2.txt};
                \end{axis}
            \end{tikzpicture}
            {\scriptsize supervised DMD\\[-4pt]\citep{Fujii19}}
        \end{minipage}
        \hspace{3mm}
        \begin{minipage}[t]{38mm}\vspace*{0pt}\centering
            \includegraphics[clip,width=15mm]{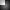}
            \hspace{0.5mm}
            \includegraphics[clip,width=15mm]{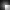}
            \par\vspace*{4mm}
            \begin{tikzpicture}[inner sep=0pt, outer sep=0pt]
                \pgfplotsset{width=48mm}
                \begin{axis}[xticklabels={,,},yticklabels={,,}]
                    \addplot table [only marks] {expt_image/disc_a0.60_cmds_1.txt};
                    \addplot table [only marks] {expt_image/disc_a0.60_cmds_2.txt};
                \end{axis}
            \end{tikzpicture}
            {\scriptsize discriminant DMD\\[-4pt](ours)}
        \end{minipage}
        \label{fig:example:mode}
    }
    \caption{(a) True dynamic modes used to generate data as \eqref{eq:synthdata}. (b) Values of the two objective functions achieved by the optimization with different $\alpha$'s. (c, upper row) Averages of the computed dynamic modes for each label. (c, lower row) Visualization of episodes using classical multidimensional scaling with the distance function induced from $k_\text{DMS}$. Each point corresponds to each episode. Blue circles and red squares denote $y=1$ and $y=2$, respectively.}
    \label{fig:example}
\end{figure*}
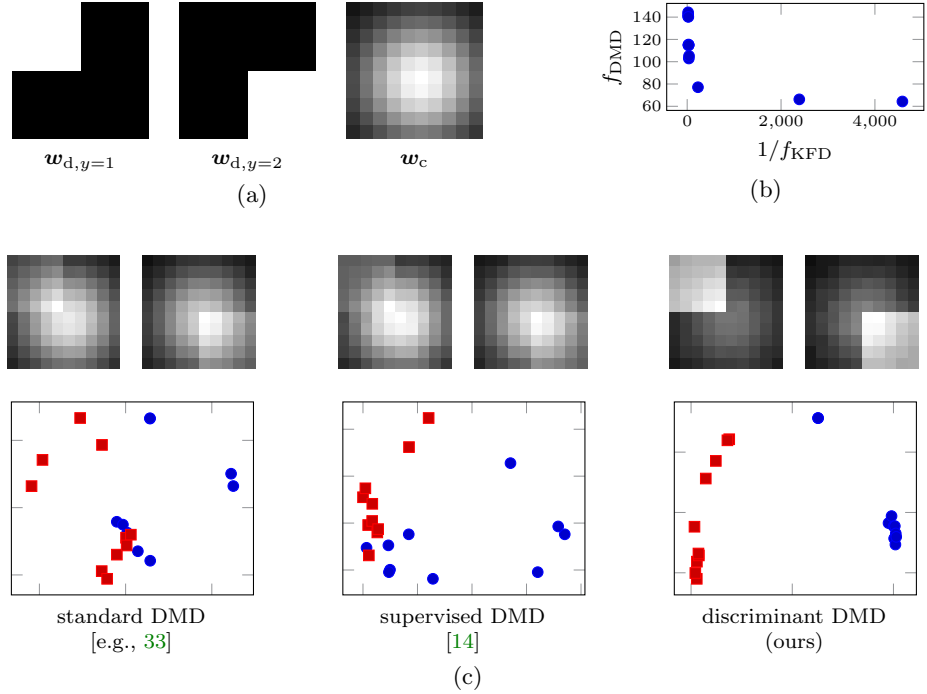

We show a numerical example with synthetic data.

\paragraph{Data Generation}
We use a synthetic dataset comprising $n=20$ episodes of length $\tau=100$, each of which is generated as
\begin{equation}\label{eq:synthdata}
    \bm{x}_{i,t} = \lambda_{\mathrm{d},i}^t \bm{w}_{\mathrm{d},y_i} + \lambda_{\mathrm{c},i}^t \bm{w}_\mathrm{c} + \bm{e}_{i,t} \in \mathbb{C}^p,
    \quad\text{for} \quad i=1,\dots,n \quad \text{and} \quad t=1,\dots,\tau,
\end{equation}
where the $i$-th episode is defined as $(\bm{x}_{i,1},\dots,\bm{x}_{i,\tau})$, and $y_i\in\{1,2\}$ is the label of the $i$-th episode.
We set $y_i=1$ for $i=1,\dots,10$ and $y_i=2$ for $i=11,\dots,20$.
\Cref{eq:synthdata} says that each episode is generated by two dynamic modes and noise.
One of the true dynamic modes, $\bm{w}_{\mathrm{d},y} \in \mathbb{R}^p$, depends on label $y$, whereas the other does not.
Our interest is to extract this $\bm{w}_{\mathrm{d},y}$ as a \emph{distinctive} dynamic mode.
In this example, we defined $\bm{w}_{\mathrm{d},y=1}$ and $\bm{w}_{\mathrm{d},y=2}$ as the left two images shown in \cref{fig:example:data}; each image is of size $10 \times 10$, and thus $p=100$.
The corresponding time-evolution parameter, $\lambda_{\mathrm{d},i} = \exp (-\gamma_\mathrm{d} + \sqrt{-1} \omega_{\mathrm{d},i}) \in \mathbb{C}$ was randomly generated for each episode by drawing $\omega_{\mathrm{d},i}$ from the uniform distribution on $[0,1]$ with $\gamma_\mathrm{d}=0.1$.
Vector $\bm{w}_{\mathrm{c}}$ is a non-distinctive dynamic mode used commonly for both $y=1$ and $y=2$ (defined as a Gaussian-like image shown in \cref{fig:example:data}), and the corresponding time-evolution parameter, $\lambda_{\mathrm{c},i}$, was randomly generated for each episode similarly to $\lambda_{\mathrm{d},i}$.
Finally, $\bm{e}\in\mathbb{C}^p$ denotes a noise vector whose elements were independently drawn from the zero-mean circularly-symmetric complex normal distribution with standard deviation $0.05$.

\paragraph{Results}
We applied the eigendecomposition-based standard DMD algorithm \citep{Tu14}, supervised DMD (\citep{Fujii19}; we review it in \cref{related}), and the proposed discriminant DMD to the synthetic dataset.
Each method was set so that it would compute a dynamic mode (i.e., $r=1$)\footnote{The true value of $r$ is obviously $2$, but we set $r=1$. We intentionally misspecified $r$ because if $r$ had been set to a true value, the problem would have been too easy.}.
Let $\hat{\bm{w}}_i$ be the dynamic mode computed for the $i$-th episode.
In \cref{fig:example:mode}, we show the averages of the dynamic modes computed by the three methods for each label (i.e., $\frac1{10} \sum_{i=1}^{10} \hat{\bm{w}}_i$ and $\frac1{10} \sum_{i=11}^{20} \hat{\bm{w}}_i$).
We can observe that the modes computed by the discriminant DMD are less occluded by the non-distinctive mode.
Moreover, we visualize the relation between the 20 episodes using the classical multidimensional scaling (MDS) with the distance function induced from $k_\text{DMS}$.
We can see that the episodes with different labels are well separated in the feature space induced from $k_\text{DMS}$.
Furthermore, we show the values of the two objectives of the discriminant DMD, $f_\text{DMD}$ and $f_\text{KFD}$, achieved by solving \cref{eq:problem} with different $\alpha$'s ($\alpha=0, 0.2, 0.4, \dots, 1.2$), in \cref{fig:example:loss}.
We can observe that the value of $f_\text{KFD}$ improves (i.e., increases), trading off the reconstruction error, $f_\text{DMD}$.



\section{Related Work}
\label{related}

\subsection{Extraction of Distinctive Patterns}
\label{related:extraction}


Informative data patterns are often obtained as a by-product of dimensionality reduction, and supervised dimensionality reduction has been studied in various contexts.
Representative examples include Fisher discriminant \citep{Fisher36,Mika99}, common spatial patterns \citep{Ramoser00}, partial least squares \citep{Wold01}, neighborhood component analysis \citep{Goldberger05}, sufficient dimensionality reduction \citep{Fukumizu09}, and supervised PCA \citep{Barshan11}.
These methods are promising tools for extracting informative patterns from labeled non-sequential data, but are not appropriate for our problem, where each data point is a sequence with a label.

Dimensionality reduction on collections of labeled sequences was addressed in the line of studies by Su \emph{et al.} \citep{Su17,Su18a,Su18b}, in which they compute discriminative ordered templates of multivariate time-series.
However, such templates do not necessarily match what we would like to compute in this work, that is, spatio-temporal coherent patterns.

Supervised DMD \citep{Fujii19} is seemingly very similar but is fundamentally different from our method in its purpose and technique.
First, whereas supervised DMD aims to extract \emph{common} dynamic modes shared by episodes of same labels, our method extracts \emph{distinctive} dynamic modes that well distinguish episodes of different labels.
More importantly, there is a critical technical difference; whereas supervised DMD modifies coefficients multiplied to dynamic modes (and thus dynamic modes remain unchanged), our method directly modifies dynamic modes (via eigenvalues) in accordance with label information.
In other words, supervised DMD adapts the \emph{usage} of dynamic modes to label information, whilst our method adapts the \emph{shape} of dynamic modes itself.



\subsection{Spectral Analysis of Koopman Operator}
\label{related:selection}

Recall that the Koopman operator is infinite-dimensional in general, and consequently, the corresponding modal decomposition is taken into infinitely-many terms (see \cref{eq:kmd1} and \cref{eq:kmd2}).
DMD is sometimes regarded as a ``finite-dimensional approximation'' of such a decomposition in some sense, but it is not straightforward to characterize which $r$ modes are actually targeted by DMD out of the infinite number of modes of the Koopman operator.
Consequently, there is no definitive way of determining the number of modes computed by DMD, $r$, nor choosing good (in some sense) dynamic modes out of computed ones.

A working practice is to determine $r$ by the effective rank of a dataset via SVD and then choose representative modes according to their energy \citep{Rowley09,Schmid10,Tu14}.
A more systematic approach is to consider additional coefficients multiplied to dynamics modes and to optimize them with sparsity regularization \citep{Jovanovic14,Fujii19} so that we can choose a relatively small number of modes effectively used in the decomposition.

Apart from DMD-like methods, other types of approaches to approximating the Koopman operator also equip strategy to pick up specific spectral components.
A method called generalized Laplace analysis \citep{Budisic12} is a rigorous way to approximate Koopman modes.
It is inherently spared for the need of mode selection because we give, in advance, eigenvalues for which modes are to be computed.
Giannakis \citep{Giannakis19} and Das and Giannakis \citep{das_delay-coordinate_2019} suggest computing the spectra of Koopman operators by a Galerkin method on eigenspaces of the Laplace--Beltrami operator of data space.
In this method, they refer to the Dirichlet energy of eigenfunctions of Koopman operator, which characterize their ``roughness,'' to select valid eigenvalues.

Given the various approaches to selecting good spectral components, we believe that our method in this work provides a new insight to this end.
That is, the proposed method is informed from label information for modifying (not selecting, though) eigenvalues and dynamic modes.
We have not rigorously characterized how the proposed method has connection to the Koopman operator theory yet, but this is an interesting research direction.


\section{Experiments}
\label{expt}

We show the application of the discriminant DMD to four real-world datasets.
The first two experiments in this section are mainly for understanding the behavior of the proposed method; we observe that the method can extract patterns that well reconstruct data \emph{and} distinguish different labels.
The last two experiments are rather for demonstrating other possible application scenes of the proposed method.


\subsection{House Temperature Data}
\label{expt:house}

\subsubsection{Dataset}

We applied the proposed discriminant DMD to measurements of temperature in a house, which exhibit particular yet varying patterns each day.
From the original dataset\footnote{\texttt{github.com/LuisM78/Appliances-energy-prediction-data}} \citep{Candanedo17}, we extracted the temperature measured in the eight rooms (kitchen, living room, laundry room, office room, bathroom, ironing room, teenager room, and parent room) of a house and at the nearby weather station.
We subtracted the measurements at the weather station from ones in the eight rooms, so a snapshot is $p=8$ dimensional.
We partitioned the original sequence by day into $n=136$ episodes (i.e., $136$ days).
We performed 6-point moving average and subsampling by 1/3 on the original data, so each episode is of length $\tau=48$ with the measurement interval being $30$ minutes.
We labeled each episode as $y=\text{``weekday''}$ (i.e., neither weekend nor holiday) or $y=\text{``holiday''}$ (i.e., Saturdays, Sundays, and national holidays).
There were $n_\text{weekday}=94$ weekdays and $n_\text{holiday}=42$ holidays.

\subsubsection{Configuration}

We applied the discriminant DMD with $r$ (the number of dynamic modes) being $r=6$ and $\alpha$ (the hyperparameter that balances reconstruction and class separation) varying from $\alpha=0$ to $\alpha=1$.
Setting $\alpha=0$ corresponds to the standard optimized DMD algorithm \citep{Chen12}, whereas $\alpha>0$ realizes our proposal for incorporating label information.
As a further baseline, we also show the results obtained by PCA on each sequence.

\clearpage
\begin{figure}[p]
    \centering
    \begin{minipage}[t]{\linewidth}
        \scriptsize\bf
        \hspace{25mm}
        MDS
        \hspace{25mm}
        All DMD eigenvalues
        \hspace{12mm}
        Median of dominant modes
    \end{minipage}
    \begin{minipage}[t]{\linewidth}
        \vspace*{0pt}
        \hspace{4mm}
        \begin{tikzpicture}[inner sep=0pt, outer sep=0pt]
            \node [rotate=90,text centered,inner sep=0pt,outer sep=0pt,minimum width=3cm] {\scriptsize PCA (w/ 2 PCs)};
        \end{tikzpicture}
        \hspace{4mm}
        \begin{tikzpicture}[inner sep=1pt, outer sep=0pt]
            \pgfplotsset{width=4.3cm,height=4.3cm}
            \begin{axis}[]
                \addplot [only marks,mark=*,color=blue] table {expt_house/raw_r3_cmds_1.txt};
                \addplot [only marks,mark=triangle*,color=red] table {expt_house/raw_r3_cmds_2.txt};
            \end{axis}
        \end{tikzpicture}
        \hspace{12mm}
        \begin{tikzpicture}[inner sep=0pt, outer sep=0pt]
            \pgfplotsset{width=3.5cm,height=2.3cm}
            \begin{axis}[ticks=none,xticklabels={,,},yticklabels={,,}]
                \addplot [only marks,mark=*,color=blue] coordinates {(0,0)};
                \addplot [only marks,mark=o,color=blue] coordinates {(1,0)};
                \node at (axis cs: 2,0) [anchor=west] {\scriptsize weekday};
                \addplot [only marks,mark=triangle*,color=red] coordinates {(0,-1)};
                \addplot [only marks,mark=triangle,color=red] coordinates {(1,-1)};
                \node at (axis cs: 2,-1) [anchor=west] {\scriptsize holiday};
                \addplot [only marks,mark=|,mark size=0,color=white] coordinates{(6.3,0.5)};
                \addplot [only marks,mark=|,mark size=0,color=white] coordinates{(-0.1,-0.5)};
                \addplot [only marks,mark=|,mark size=0,color=white] coordinates{(6.3,-1.5)};
            \end{axis}
        \end{tikzpicture}
        \hspace{1.2cm}
        \hspace{12mm}
        \includegraphics[clip,height=3cm,trim=0 22 66 11]{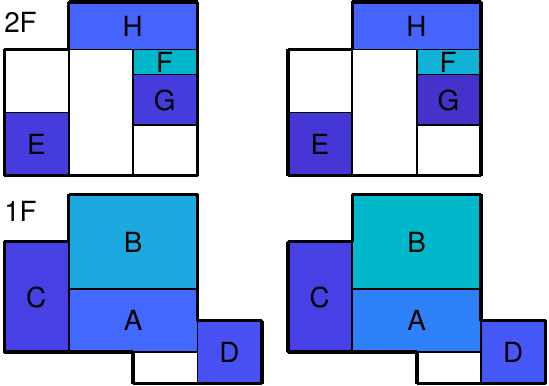}
    \end{minipage}
    \par\vspace*{12pt}
    \begin{minipage}[t]{\linewidth}
        \vspace*{0pt}
        \hspace{4mm}
        \begin{tikzpicture}[inner sep=0pt, outer sep=0pt]
            \node [rotate=90,text centered,inner sep=0pt,outer sep=0pt,minimum width=3cm] {\scriptsize DMD ($\alpha=0$)};
        \end{tikzpicture}
        \hspace{4mm}
        \begin{tikzpicture}[inner sep=1pt, outer sep=0pt]
            \pgfplotsset{width=4.3cm,height=4.3cm}
            \begin{axis}[]
                \addplot [only marks,mark=*,color=blue] table {expt_house/dmd_a0.0_cmds_1.txt};
                \addplot [only marks,mark=triangle*,color=red] table {expt_house/dmd_a0.0_cmds_2.txt};
            \end{axis}
        \end{tikzpicture}
        \hspace{12mm}
        \begin{tikzpicture}[inner sep=1pt, outer sep=0pt]
            \pgfplotsset{width=3.9cm,height=3.9cm}
            \begin{axis}[xlabel={$\operatorname{Re}(\log\lambda)/(2\pi\Delta t)$},
                         ylabel={$\operatorname{Im}(\log\lambda)/(2\pi\Delta t)$},
                         xmin=-0.3,xmax=0.15,ymin=-0.2,ymax=0.2]
                \addplot [only marks,mark=o,color=blue] table {expt_house/dmd_a0.0_eigval_1.txt};
                \addplot [only marks,mark=triangle,color=red] table {expt_house/dmd_a0.0_eigval_2.txt};
            \end{axis}
        \end{tikzpicture}
        \hspace{12mm}
        \includegraphics[clip,height=3cm,trim=0 22 66 11]{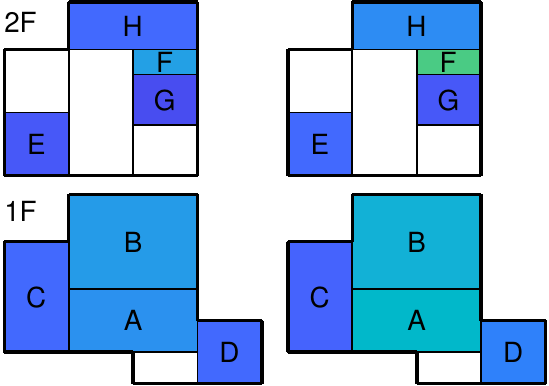}
    \end{minipage}
    \par\vspace*{12pt}
    \begin{minipage}[t]{\linewidth}
        \vspace*{0pt}
        \hspace{4mm}
        \begin{tikzpicture}[inner sep=0pt, outer sep=0pt]
            \node [rotate=90,text centered,inner sep=0pt,outer sep=0pt,minimum width=3cm] {\scriptsize discDMD ($\alpha=0.2$)};
        \end{tikzpicture}
        \hspace{4mm}
        \begin{tikzpicture}[inner sep=1pt, outer sep=0pt]
            \pgfplotsset{width=4.3cm,height=4.3cm}
            \begin{axis}[]
                \addplot [only marks,mark=*,color=blue] table {expt_house/dmd_a0.2_cmds_1.txt};
                \addplot [only marks,mark=triangle*,color=red] table {expt_house/dmd_a0.2_cmds_2.txt};
            \end{axis}
        \end{tikzpicture}
        \hspace{12mm}
        \begin{tikzpicture}[inner sep=1pt, outer sep=0pt]
            \pgfplotsset{width=3.9cm,height=3.9cm}
            \begin{axis}[xlabel={$\operatorname{Re}(\log\lambda)/(2\pi\Delta t)$},
                         ylabel={$\operatorname{Im}(\log\lambda)/(2\pi\Delta t)$},
                         xmin=-0.3,xmax=0.15,ymin=-0.2,ymax=0.2]
                \addplot [only marks,mark=o,color=blue] table {expt_house/dmd_a0.2_eigval_1.txt};
                \addplot [only marks,mark=triangle,color=red] table {expt_house/dmd_a0.2_eigval_2.txt};
            \end{axis}
        \end{tikzpicture}
        \hspace{12mm}
        \includegraphics[clip,height=3cm,trim=0 22 66 11]{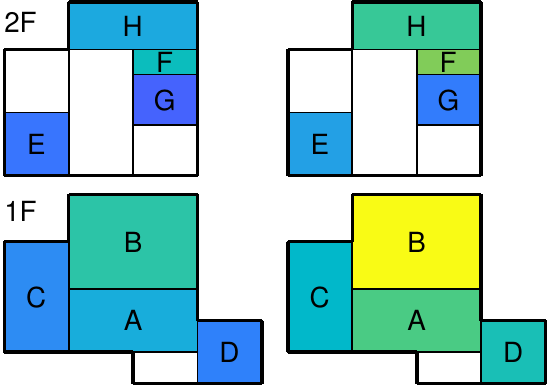}
    \end{minipage}
    \par\vspace*{12pt}
    \begin{minipage}[t]{\linewidth}
        \vspace*{0pt}
        \hspace{4mm}
        \begin{tikzpicture}[inner sep=0pt, outer sep=0pt]
            \node [rotate=90,text centered,inner sep=0pt,outer sep=0pt,minimum width=3cm] {\scriptsize discDMD ($\alpha=1.0$)};
        \end{tikzpicture}
        \hspace{4mm}
        \begin{tikzpicture}[inner sep=1pt, outer sep=0pt]
            \pgfplotsset{width=4.3cm,height=4.3cm}
            \begin{axis}[]
                \addplot [only marks,mark=*,color=blue] table {expt_house/dmd_a1.0_cmds_1.txt};
                \addplot [only marks,mark=triangle*,color=red] table {expt_house/dmd_a1.0_cmds_2.txt};
            \end{axis}
        \end{tikzpicture}
        \hspace{12mm}
        \begin{tikzpicture}[inner sep=1pt, outer sep=0pt]
            \pgfplotsset{width=3.9cm,height=3.9cm}
            \begin{axis}[xlabel={$\operatorname{Re}(\log\lambda)/(2\pi\Delta t)$},
                         ylabel={$\operatorname{Im}(\log\lambda)/(2\pi\Delta t)$},
                         xmin=-0.3,xmax=0.15,ymin=-0.2,ymax=0.2]
                \addplot [only marks,mark=o,color=blue] table {expt_house/dmd_a1.0_eigval_1.txt};
                \addplot [only marks,mark=triangle,color=red] table {expt_house/dmd_a1.0_eigval_2.txt};
            \end{axis}
        \end{tikzpicture}
        \hspace{12mm}
        \includegraphics[clip,height=3cm,trim=0 22 66 11]{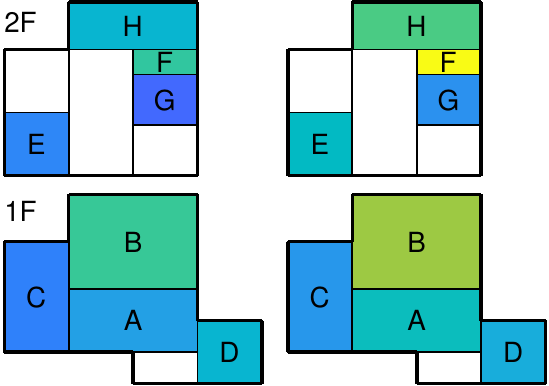}
    \end{minipage}
    \par\vspace*{5pt}
    \begin{minipage}[t]{\linewidth}
        \vspace*{0pt}
        \hspace{106mm}
        \begin{tikzpicture}[inner sep=0pt, outer sep=0pt]
            \node [text centered,anchor=south] at (-1.1,0.5) {\scriptsize weekday};
            \node [text centered,anchor=south] at (1.1,0.5) {\scriptsize holiday};
            \node [draw,anchor=south] at (0,0) {\includegraphics[clip,height=3mm,width=40mm]{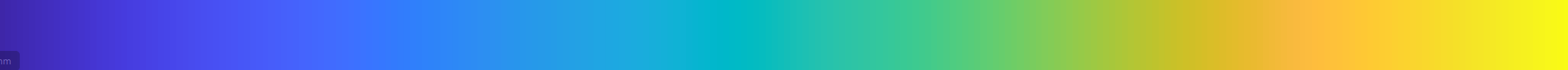}};
            \node [text centered,anchor=north] at (-2,-0.1) {\tiny $0.0$};
            \node [text centered,anchor=north] at (2,-0.1) {\tiny $1.2$};
        \end{tikzpicture}
    \end{minipage}
    \caption{Results on the house temperature dataset. (\emph{left}) Visualization of episodes via MDS with the projection kernel. Each point corresponds to each episode of the dataset. (\emph{center}) DMD eigenvalues. All the eigenvalues computed from all episodes are plotted altogether. (\emph{right}) Visualization of average dominant modes overlaid on the floor plan. The letters denote: (A) kitchen, (B) living room, (C) laundry room, (D) office room, (E) bathroom, (F) ironing room, (G) teenager room, and (H) parent room. Best viewed in color.}
    \label{fig:house}
\end{figure}
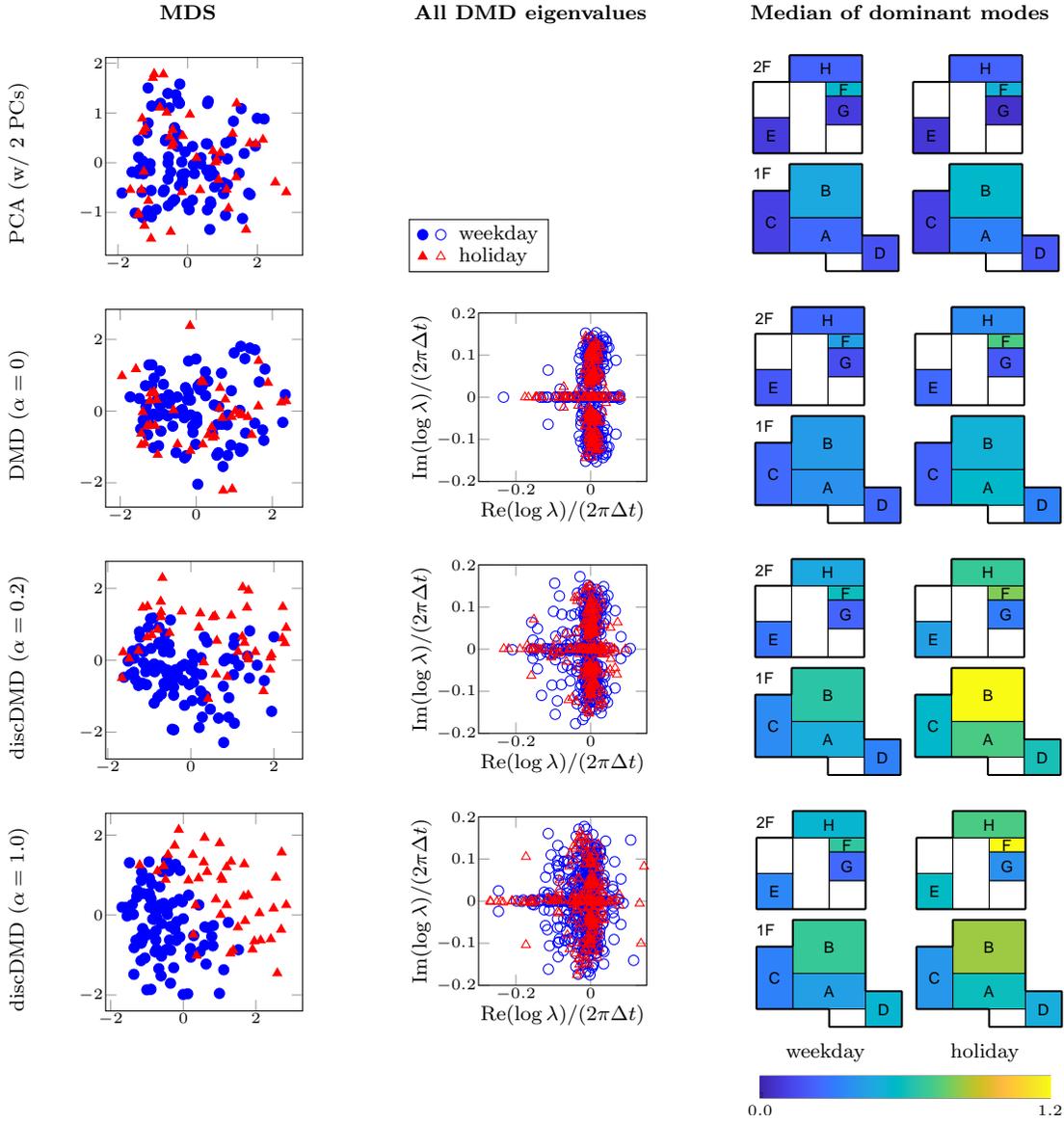

\clearpage

\subsubsection{Results}

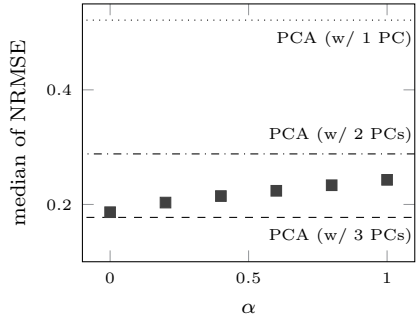
\begin{figure}[t]
    \centering
    \begin{tikzpicture}[inner sep=4pt, outer sep=0pt]
        \pgfplotsset{width=6cm,height=5cm}
        \begin{axis}[
            xticklabel style={
                /pgf/number format/fixed,
                /pgf/number format/precision=2
            },
            xmin=-0.1, xmax=1.1, ymax=0.55, ymin=0.1,
            scaled x ticks=false,
            ylabel near ticks,
            xlabel near ticks,
            xlabel={$\alpha$},
            ylabel={median of NRMSE},
            label style={font=\scriptsize},
        ]
            \addplot [mark=square*, color=darkgray, only marks] table [x=alpha, y=median_nrmse] {expt_house/dmd_loss.txt};
            \addplot[mark=none,black,dotted] coordinates {(-0.2,0.5217) (1.2,0.5217)};
            \node at (axis cs: 1.13,0.5217) [anchor=north east] {\tiny PCA (w/ 1 PC)};
            \addplot[mark=none,black,dash dot] coordinates {(-0.2,0.2882) (1.2,0.2882)};
            \node at (axis cs: 1.13,0.2882) [anchor=south east] {\tiny PCA (w/ 2 PCs)};
            \addplot[mark=none,black,dashed] coordinates {(-0.2,0.1774) (1.2,0.1774)};
            \node at (axis cs: 1.13,0.1774) [anchor=north east] {\tiny PCA (w/ 3 PCs)};
        \end{axis}
    \end{tikzpicture}
    \caption{Median of NRMSE of the reconstruction by the discriminant DMD with different values of $\alpha$, for the house temperature dataset. $\alpha=0$ corresponds to the standard DMD, and $\alpha>0$ corresponds to the discriminant DMD. The reconstruction errors by PCA with one, two, or three PCs are also shown for comparison.}
    \label{fig:house:loss}
\end{figure}

In the left column of \cref{fig:house}, we show the two-dimensional embedding of the episodes by MDS with the distance between episodes computed via the projection kernel.
The projection kernel was computed between sets of dynamic modes for DMDs and sets of principal components (PCs) for PCA\footnote{Projection kernel on PCs is similar to its original usage \citep{Hamm08}.}.
We can observe that the episodes with different labels are well separated with the discriminant DMD ($\alpha>0$) as expected from the formulation.

In the center column of \cref{fig:house}, we plot the values of $\lambda$ computed on all the episodes.
This quantity, $\lambda$, appears both in standard DMD and our discriminant DMD (see \cref{eq:dmd} and \cref{eq:optdmdprob}), and we refer to it in both cases simply as DMD eigenvalues.
The DMD eigenvalues of the two labels distribute quite similarly when $\alpha=0$ (i.e., no label information is reflected), whereas the distributions are distinctively different between the two labels when $\alpha>0$.

In the right column of \cref{fig:house}, we visualize the dynamic modes\footnote{As we compute multiple dynamic modes for all the episodes, we examined summary statistics of the dynamic modes for each label.
First, we chose the dominant mode with the largest norm $\Vert \bm{w} \Vert$ for each episode.
Then, we calculated the element-wise median of the dominant modes over all the episodes for each label.}
by painting each room of a floor plan of the house according to the dynamic modes.
With the discriminant DMD (i.e., $\alpha>0$), the difference between $y=\text{``weekday''}$ and $y=\text{``holiday''}$ are consistently emphasized in some rooms such as B (living room) and F (ironing room).
We can utilize such information, for example, for designing a controller of air conditioning systems.

In \cref{fig:house:loss}, we show the median of the normalized root mean square error (NRMSE) between the original time-series sequences and the reconstructed ones by the discriminant DMD with different $\alpha$ as well as by PCA.
We can observe that the reconstruction by the discriminant DMD achieves the similar reconstruction errors as PCA with two or three PCs, regardless of the value of $\alpha$.
Recall that we set $r=6$ for the discriminant DMD.
Because a DMD-like decomposition usually results in (nearly) complex conjugate pairs of modes for real-valued data, it is reasonable to expect that the reconstruction capability of DMD-like decomposition with $r=6$ modes (i.e., three pairs) roughly corresponds to that of PCA with three PCs.
The result of $\alpha=0$ (i.e., standard DMD) in \cref{fig:house:loss} fits this expectation almost completely.
Moreover, we can see the reconstruction capability with $\alpha>0$ (i.e., discriminant DMD) is not severely sacrificed, which supports the validity of the proposed method in the sense that it can compute distinctive \emph{and} representative dynamic modes.


\subsection{Motion Capture Data}
\label{expt:mocap}

\subsubsection{Dataset}

\begin{figure}[t]
    \centering
    \includegraphics[clip,height=4cm]{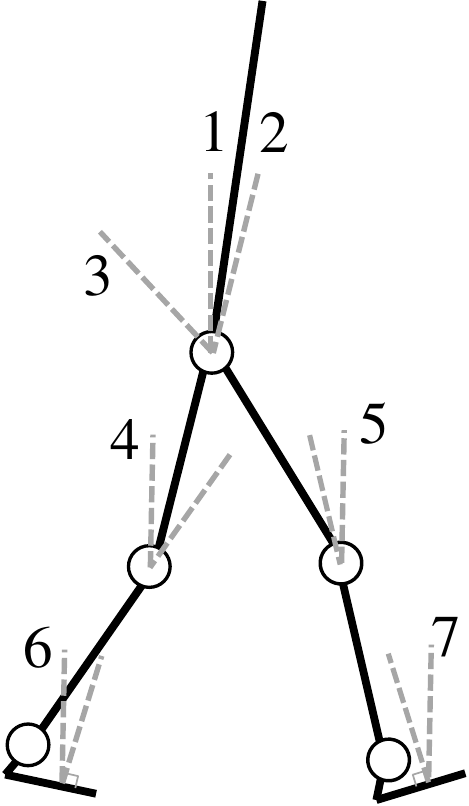}
    \caption{Seven angles used as features of the motion capture data.}
    \label{fig:mocap:stick}
\end{figure}

We also applied the discriminant DMD to measurements of human locomotion, which has already been a target of analysis of standard DMD techniques \citep{Fujii19locom}.
We prepared a dataset from the CMU Graphics Lab Motion Capture Database\footnote{\url{mocap.cs.cmu.edu}}.
We chose $n_\text{walk}=40$ episodes of ``walk'' motion and $n_\text{run/jog}=40$ episodes of ``run/jog'' motion and constructed a dataset with these $n=80$ episodes.
We only used the first $\tau=128$ timesteps (about one second) of each sequence.
We transformed the original data into the angles of trunk, thigh, shank, and foot elevation in the sagittal plane (schematically shown in \cref{fig:mocap:stick}) for extracting intersegmental motion coherence \citep{Funato10}.
Consequently, each episode is a $p=7$ dimensional time-series sequence.
As preprocessing, we normalized each feature and applied a low-pass filter at $6$ [Hz].

\subsubsection{Configuration}

The configuration is similar to that in the previous experiment; we applied the discriminant DMD with $r=4$, and the hyperparameter $\alpha$ was varied from $\alpha=0$ to $\alpha=1$.
We also show the results obtained by PCA on each sequence.

\subsubsection{Results}

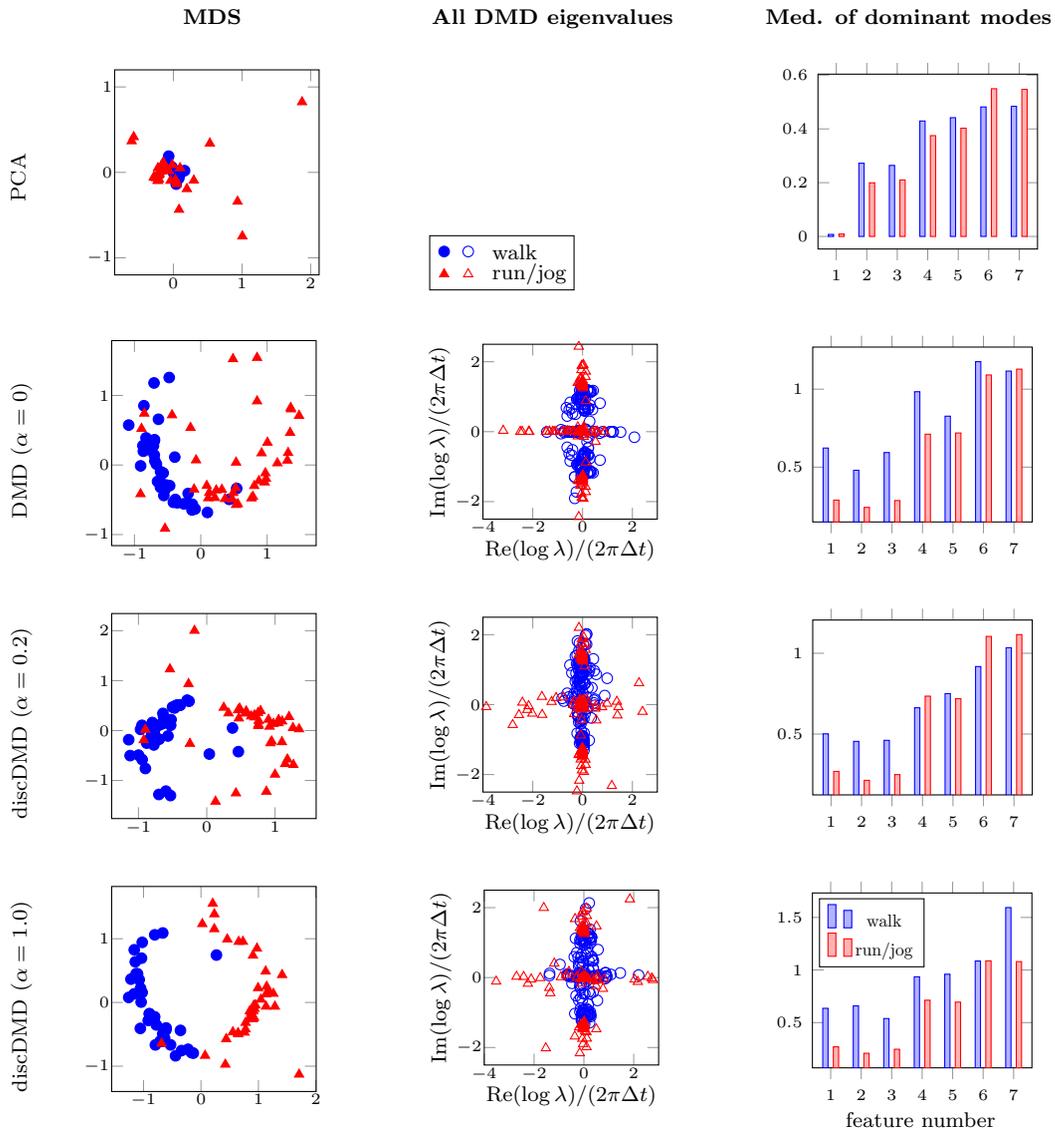
\begin{figure}[t]
    \centering
    \begin{minipage}[t]{\linewidth}
        \scriptsize\bf
        \hspace{31mm}
        MDS
        \hspace{23mm}
        All DMD eigenvalues
        \hspace{10mm}
        Med. of dominant modes
    \end{minipage}
    \begin{minipage}[t]{\linewidth}
        \vspace*{0pt}
        \hspace{8mm}
        \begin{tikzpicture}[inner sep=0pt, outer sep=0pt]
            \node [rotate=90,text centered,inner sep=0pt,outer sep=0pt,minimum width=3cm] {\scriptsize PCA};
        \end{tikzpicture}
        \hspace{4mm}
        \hspace{0pt}
        \begin{tikzpicture}[inner sep=1pt, outer sep=0pt]
            \pgfplotsset{width=4.3cm,height=4.3cm}
            \begin{axis}[ymin=-1.2, ymax=1.2]
                \addplot [only marks,mark=*,color=blue] table {expt_mocap/raw_r2_cmds_1.txt};
                \addplot [only marks,mark=triangle*,color=red] table {expt_mocap/raw_r2_cmds_2.txt};
            \end{axis}
        \end{tikzpicture}
        \hspace{12mm}
        \begin{tikzpicture}[inner sep=0pt, outer sep=0pt]
            \pgfplotsset{width=3.5cm,height=2.3cm}
            \begin{axis}[ticks=none,xticklabels={,,},yticklabels={,,}]
                \addplot [only marks,mark=*,color=blue] coordinates {(0,0)};
                \addplot [only marks,mark=o,color=blue] coordinates {(1,0)};
                \node at (axis cs: 2,0) [anchor=west] {\scriptsize walk};
                \addplot [only marks,mark=triangle*,color=red] coordinates {(0,-1)};
                \addplot [only marks,mark=triangle,color=red] coordinates {(1,-1)};
                \node at (axis cs: 2,-1) [anchor=west] {\scriptsize run/jog};
                \addplot [only marks,mark=|,mark size=0,color=white] coordinates{(4.9,0.5)};
                \addplot [only marks,mark=|,mark size=0,color=white] coordinates{(-0.1,-0.5)};
                \addplot [only marks,mark=|,mark size=0,color=white] coordinates{(4.9,-1.5)};
            \end{axis}
        \end{tikzpicture}
        \hspace{1.05cm}
        \hspace{12mm}
        \begin{tikzpicture}
            \pgfplotsset{width=4.5cm,height=3.9cm}
            \begin{axis}[ybar,bar width=2pt,xtick=data,xticklabels={1,2,3,4,5,6,7}]
                \addplot table [x=feat, y=avgmode1] {expt_mocap/pca_mode.txt};
                \addplot table [x=feat, y=avgmode2] {expt_mocap/pca_mode.txt};    
            \end{axis}
        \end{tikzpicture}
    \end{minipage}
    \par\vspace*{16pt}
    \begin{minipage}[t]{\linewidth}
        \vspace*{0pt}
        \hspace{8mm}
        \begin{tikzpicture}[inner sep=0pt, outer sep=0pt]
            \node [rotate=90,text centered,inner sep=0pt,outer sep=0pt,minimum width=3cm] {\scriptsize DMD ($\alpha=0$)};
        \end{tikzpicture}
        \hspace{4mm}
        \begin{tikzpicture}[inner sep=1pt, outer sep=0pt]
            \pgfplotsset{width=4.3cm,height=4.3cm}
            \begin{axis}[]
                \addplot [only marks,mark=*,color=blue] table {expt_mocap/dmd_a0.0_cmds_1.txt};
                \addplot [only marks,mark=triangle*,color=red] table {expt_mocap/dmd_a0.0_cmds_2.txt};
            \end{axis}
        \end{tikzpicture}
        \hspace{12mm}
        \begin{tikzpicture}[inner sep=1pt, outer sep=0pt]
            \pgfplotsset{width=3.9cm,height=3.9cm}
            \begin{axis}[xlabel={$\operatorname{Re}(\log\lambda)/(2\pi\Delta t)$},
                         ylabel={$\operatorname{Im}(\log\lambda)/(2\pi\Delta t)$},
                         xmin=-4,xmax=3,ymin=-2.5,ymax=2.5]
                \addplot [only marks,mark=o,color=blue] table {expt_mocap/dmd_a0.0_eigval_1.txt};
                \addplot [only marks,mark=triangle,color=red] table {expt_mocap/dmd_a0.0_eigval_2.txt};
            \end{axis}
        \end{tikzpicture}
        \hspace{12mm}
        \begin{tikzpicture}
            \pgfplotsset{width=4.5cm,height=3.9cm}
            \begin{axis}[ybar,bar width=2pt,xtick=data,xticklabels={1,2,3,4,5,6,7}]
                \addplot table [x=feat, y=avgmode1] {expt_mocap/dmd_a0.0_mode.txt};
                \addplot table [x=feat, y=avgmode2] {expt_mocap/dmd_a0.0_mode.txt};    
            \end{axis}
        \end{tikzpicture}
    \end{minipage}
    \par\vspace*{16pt}
    \begin{minipage}[t]{\linewidth}
        \vspace*{0pt}
        \hspace{8mm}
        \begin{tikzpicture}[inner sep=0pt, outer sep=0pt]
            \node [rotate=90,text centered,inner sep=0pt,outer sep=0pt,minimum width=3cm] {\scriptsize discDMD ($\alpha=0.2$)};
        \end{tikzpicture}
        \hspace{4mm}
        \begin{tikzpicture}[inner sep=1pt, outer sep=0pt]
            \pgfplotsset{width=4.3cm,height=4.3cm}
            \begin{axis}[]
                \addplot [only marks,mark=*,color=blue] table {expt_mocap/dmd_a0.2_cmds_1.txt};
                \addplot [only marks,mark=triangle*,color=red] table {expt_mocap/dmd_a0.2_cmds_2.txt};
            \end{axis}
        \end{tikzpicture}
        \hspace{12mm}
        \begin{tikzpicture}[inner sep=1pt, outer sep=0pt]
            \pgfplotsset{width=3.9cm,height=3.9cm}
            \begin{axis}[xlabel={$\operatorname{Re}(\log\lambda)/(2\pi\Delta t)$},
                         ylabel={$\operatorname{Im}(\log\lambda)/(2\pi\Delta t)$},
                         xmin=-4,xmax=3,ymin=-2.5,ymax=2.5]
                \addplot [only marks,mark=o,color=blue] table {expt_mocap/dmd_a0.2_eigval_1.txt};
                \addplot [only marks,mark=triangle,color=red] table {expt_mocap/dmd_a0.2_eigval_2.txt};
            \end{axis}
        \end{tikzpicture}
        \hspace{12mm}
        \begin{tikzpicture}
            \pgfplotsset{width=4.5cm,height=3.9cm}
            \begin{axis}[ybar,bar width=2pt,xtick=data,xticklabels={1,2,3,4,5,6,7}]
                \addplot table [x=feat, y=avgmode1] {expt_mocap/dmd_a0.2_mode.txt};
                \addplot table [x=feat, y=avgmode2] {expt_mocap/dmd_a0.2_mode.txt};    
            \end{axis}
        \end{tikzpicture}
    \end{minipage}
    \par\vspace*{16pt}
    \begin{minipage}[t]{\linewidth}
        \vspace*{0pt}
        \hspace{8mm}
        \begin{tikzpicture}[inner sep=0pt, outer sep=0pt]
            \node [rotate=90,text centered,inner sep=0pt,outer sep=0pt,minimum width=3cm] {\scriptsize discDMD ($\alpha=1.0$)};
        \end{tikzpicture}
        \hspace{4mm}
        \begin{tikzpicture}[inner sep=1pt, outer sep=0pt]
            \pgfplotsset{width=4.3cm,height=4.3cm}
            \begin{axis}[]
                \addplot [only marks,mark=*,color=blue] table {expt_mocap/dmd_a1.0_cmds_1.txt};
                \addplot [only marks,mark=triangle*,color=red] table {expt_mocap/dmd_a1.0_cmds_2.txt};
            \end{axis}
        \end{tikzpicture}
        \hspace{12mm}
        \hspace{-2mm}
        \begin{tikzpicture}[inner sep=1pt, outer sep=0pt]
            \pgfplotsset{width=3.9cm,height=3.9cm}
            \begin{axis}[xlabel={$\operatorname{Re}(\log\lambda)/(2\pi\Delta t)$},
                         ylabel={$\operatorname{Im}(\log\lambda)/(2\pi\Delta t)$},
                         xmin=-4,xmax=3,ymin=-2.5,ymax=2.5]
                \addplot [only marks,mark=o,color=blue] table {expt_mocap/dmd_a1.0_eigval_1.txt};
                \addplot [only marks,mark=triangle,color=red] table {expt_mocap/dmd_a1.0_eigval_2.txt};
            \end{axis}
        \end{tikzpicture}
        \hspace{12mm}
        \hspace{-1.5mm}
        \begin{tikzpicture}
            \pgfplotsset{width=4.5cm,height=3.9cm}
            \begin{axis}[ybar,bar width=2pt,xtick=data,xticklabels={1,2,3,4,5,6,7},
                legend pos={north west},
                legend entries={walk,run/jog},
                legend style={nodes={scale=0.6, transform shape}}]
                \addplot table [x=feat, y=avgmode1] {expt_mocap/dmd_a1.0_mode.txt};
                \addplot table [x=feat, y=avgmode2] {expt_mocap/dmd_a1.0_mode.txt};    
            \end{axis}
        \end{tikzpicture}
    \end{minipage}
    \par\vspace*{0pt}
    \begin{minipage}[t]{\linewidth}
        \vspace*{0pt}
        \hspace{119mm}
        {\scriptsize feature number}
    \end{minipage}
    \caption{Results on the motion capture dataset. (\emph{left}) Visualization of episodes via MDS with the projection kernel. Each point corresponds to each episode of the dataset. (\emph{center}) DMD eigenvalues. All the eigenvalues computed from all episodes are plotted altogether. (\emph{right}) Visualization of average dominant modes. The feature numbers correspond to ones shown in Fig.~\ref{fig:mocap:stick}.}
    \label{fig:mocap}
\end{figure}

In the left column of \cref{fig:mocap}, we show the two-dimensional embedding of the episodes computed similarly to the previous experiment, that is, via MDS with projection kernels.
The overlap of the episodes of different labels is alleviated by the proposed method with $\alpha>0$.
In the center and right columns of \cref{fig:mocap}, we show all the DMD eigenvalues and the medians of dominant modes, respectively, similarly to the previous experiment.
The distinction between $y=\text{``walk''}$ and $y=\text{``run/jog''}$ is less obvious visually in this case, but such information, possibly combined with domain knowledge, may play an important role in gait analysis.

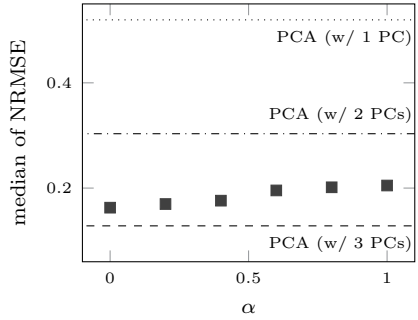
\begin{figure}[t]
    \centering
    \begin{tikzpicture}[inner sep=4pt, outer sep=0pt]
        \pgfplotsset{width=6cm,height=5cm}
        \begin{axis}[
            xticklabel style={
                /pgf/number format/fixed,
                /pgf/number format/precision=2
            },
            xmin=-0.1, xmax=1.1, ymax=0.55, ymin=0.06,
            scaled x ticks=false,
            ylabel near ticks,
            xlabel near ticks,
            xlabel={$\alpha$},
            ylabel={median of NRMSE},
            label style={font=\scriptsize},
        ]
            \addplot [mark=square*, color=darkgray, only marks] table [x=alpha, y=median_nrmse] {expt_mocap/dmd_loss.txt};
            \addplot[mark=none,black,dotted] coordinates {(-0.2,0.5197) (1.2,0.5197)};
            \node at (axis cs: 1.13,0.5197) [anchor=north east] {\tiny PCA (w/ 1 PC)};
            \addplot[mark=none,black,dash dot] coordinates {(-0.2,0.3033) (1.2,0.3033)};
            \node at (axis cs: 1.13,0.3033) [anchor=south east] {\tiny PCA (w/ 2 PCs)};
            \addplot[mark=none,black,dashed] coordinates {(-0.2,0.1282) (1.2,0.1282)};
            \node at (axis cs: 1.13,0.1282) [anchor=north east] {\tiny PCA (w/ 3 PCs)};
        \end{axis}
    \end{tikzpicture}
    \caption{Median of NRMSE of the reconstruction by the discriminant DMD with different values of $\alpha$, for the motion capture dataset. $\alpha=0$ corresponds to standard DMD, and $\alpha>0$ corresponds to the discriminant DMD. The reconstruction errors by PCA with one, two, or three PCs are also shown for comparison.}
    \label{fig:mocap:loss}
\end{figure}

In \cref{fig:mocap:loss}, we show the median of the NRMSE between the original sequences and the reconstructed ones, with which we can obtain similar observations as in the previous experiment.

\clearpage

\subsection{Taxi Trip Data}
\label{expt:taxi}

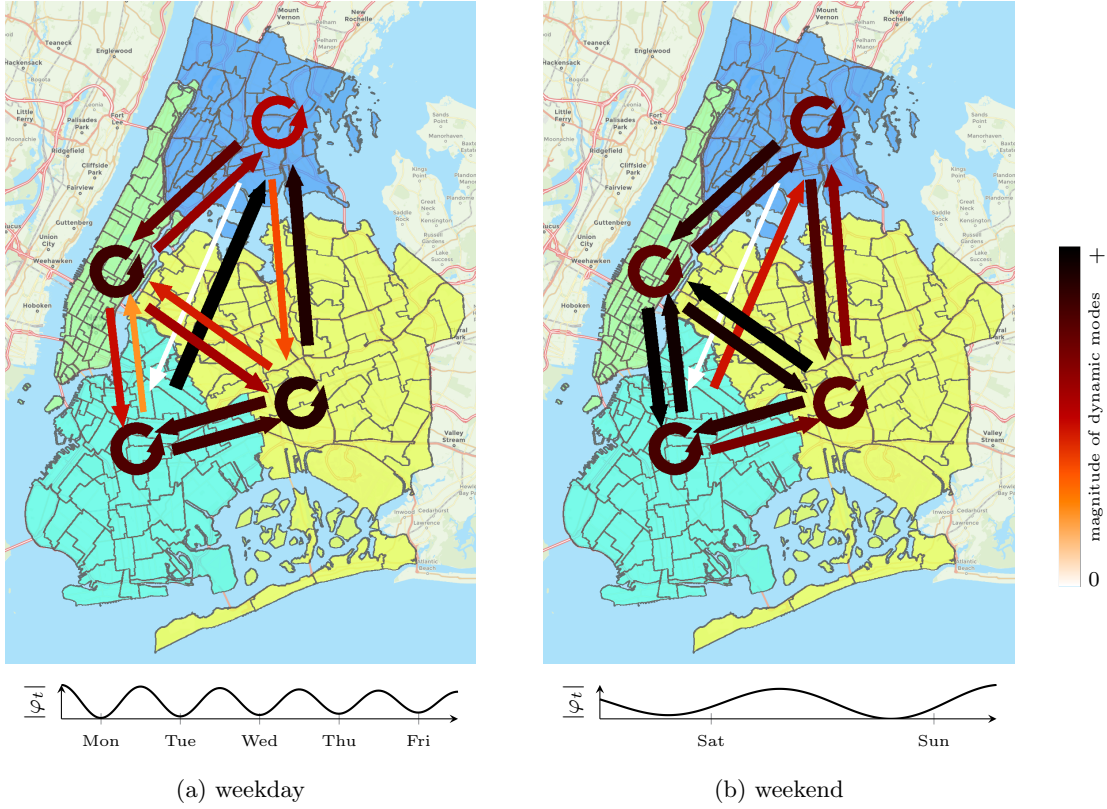
\begin{figure}[t]
    \begin{minipage}[t]{0.45\textwidth}\vspace{0pt}\centering\subfloat[weekday]{\label{fig:taxi:weekday}
        \begin{tabular}{l}
            \includegraphics[clip,width=0.89\textwidth,trim={3.5cm 3cm 4cm 7.4cm}]{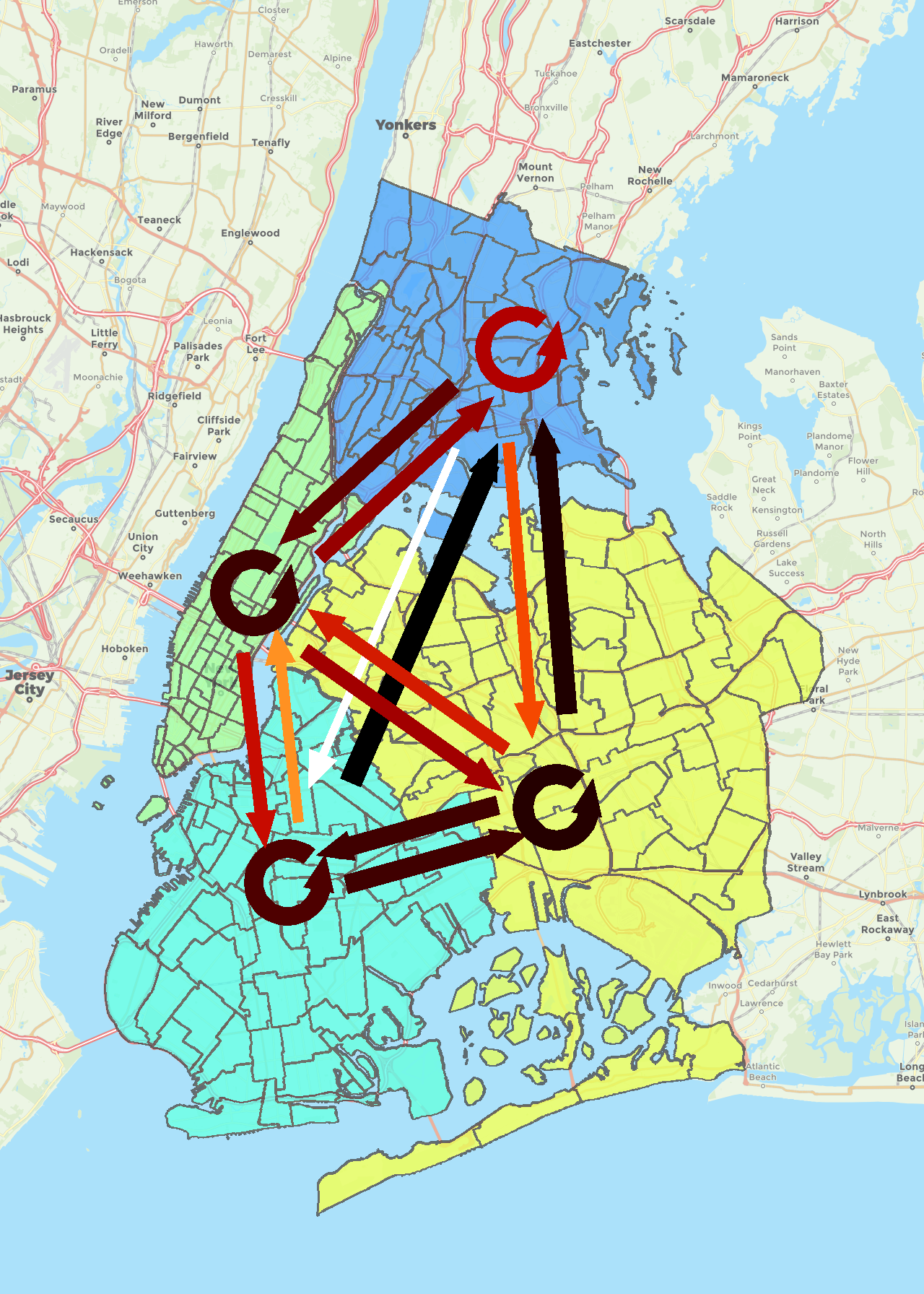} \\
            \pgfplotsset{width=0.75\textwidth, height=0.45cm}
            \begin{tikzpicture}
                \begin{axis}[
                    axis lines = left,
                    xtick = {3.14, 9.42, 15.7, 22.0, 28.3},
                    xticklabels = {Mon, Tue, Wed, Thu, Fri},
                    ymin = -1, ymax = 1,
                    ytick = \empty, yticklabels = {,,},
                    ylabel = $\vert\varphi_t\vert$,
                    font = \tiny, enlargelimits = false, scale only axis]
                    \addplot[domain=0:5*2*pi, samples=200, thick]{exp(-0.0163*x)*cos(deg(x))};
                \end{axis}
            \end{tikzpicture}
        \end{tabular}
    }\end{minipage}
    \begin{minipage}[t]{0.45\textwidth}\vspace{0pt}\centering\subfloat[weekend]{\label{fig:taxi:weekend}
        \begin{tabular}{l}
            \includegraphics[clip,width=0.89\textwidth,trim={3.5cm 3cm 4cm 7.4cm}]{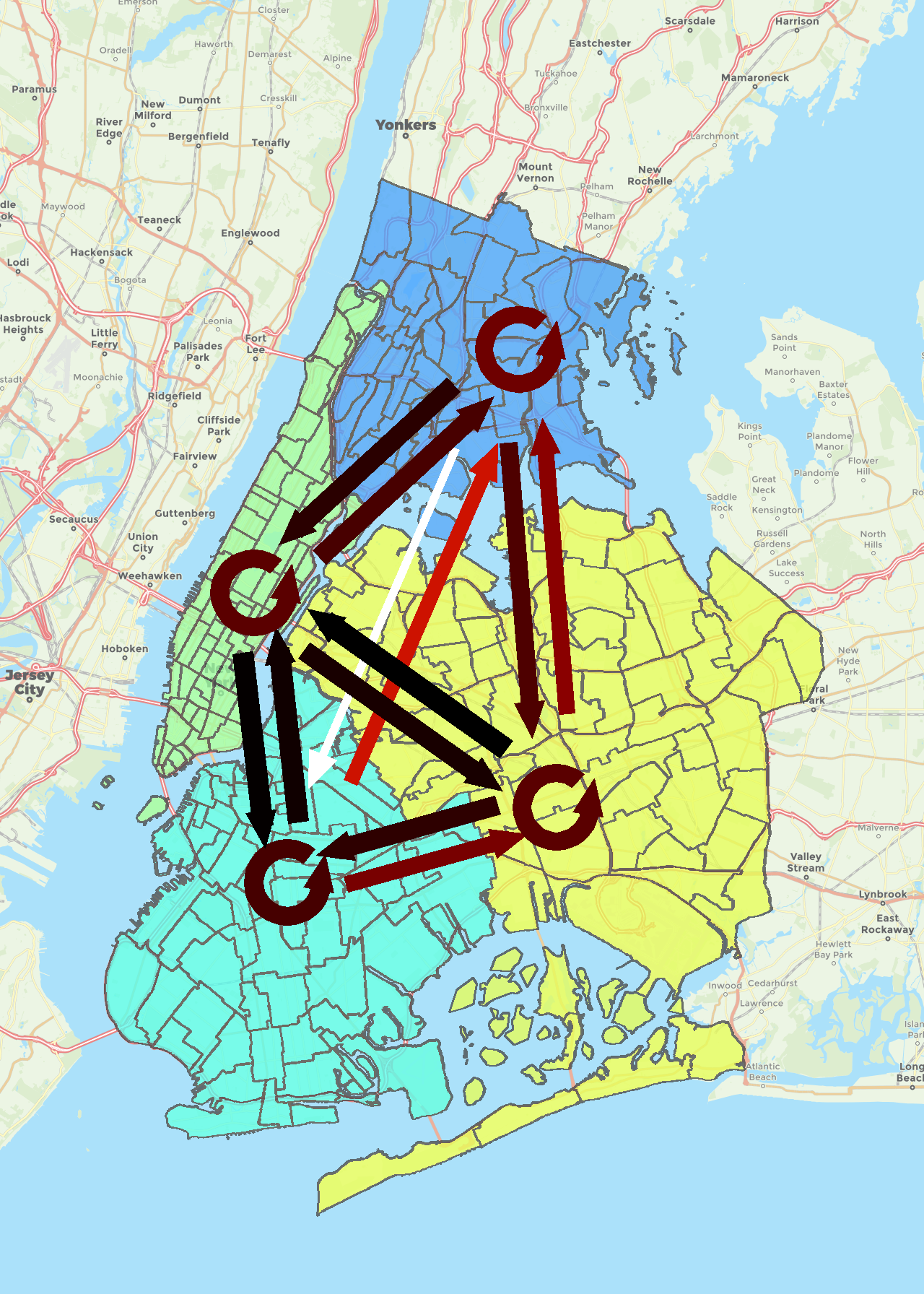} \\
            \pgfplotsset{width=0.75\textwidth, height=0.45cm}
            \begin{tikzpicture}
                \begin{axis}[
                    axis lines = left,
                    xtick = {3.14, 9.42},
                    xticklabels = {Sat, Sun},
                    ytick = \empty, yticklabels = {,,},
                    ylabel = $\vert\varphi_t\vert$,
                    font = \tiny, enlargelimits = false, scale only axis]
                    \addplot[domain=0:1.78*2*pi, samples=200, thick]{9*sin(deg(x)+deg(0.9*pi))*exp(0.0418*x)};
                \end{axis}
            \end{tikzpicture}
        \end{tabular}
    }\end{minipage}
    \begin{minipage}[t]{0.05\textwidth}\vspace{3.25cm}\centering
        \begin{tikzpicture}
            \node[draw, thin, inner sep=0pt, anchor=south east] at (0,0) {\includegraphics[height=4.5cm,width=0.27cm]{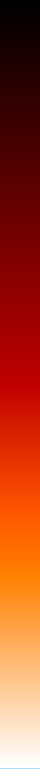}};
            \node[inner sep=0pt, anchor=south west] at (0.1,0) {\scriptsize 0};
            \node[inner sep=0pt, anchor=north west] at (0.1,4.5) {\scriptsize +};
            \node[inner sep=0pt, rotate=90] at (0.2,2.25) {\tiny magnitude of dynamic modes};
        \end{tikzpicture}
    \end{minipage}
    \caption{Examples of the distinctive coherent patterns extracted from the taxi trip data. Areas in green, light blue, yellow, and blue indicate Manhattan, Brooklyn, Queens, and Bronx, respectively. (\emph{upper}) Magnitude of the coherent patterns. (\emph{lower}) Temporal profiles of corresponding dynamics.
    }
    \label{fig:taxi}
\end{figure}

\subsubsection{Dataset}

Certain kinds of spatio-temporal data have characteristic structures we can anticipate based on knowledge of social and biological rhythms (e.g., days, months, and years).
The discriminant DMD will be remarkably useful for such data because we can examine specific coherent structures focusing on their dynamical property, that is, frequencies of rhythms.
As an example of such a usage, we applied the discriminant DMD to records of taxi trips.
The dataset is a record of the numbers of taxies that departed and arrived between four boroughs of New York City (Manhattan, Brooklyn, Queens, and Bronx)\footnote{We used trip records of boro taxies that are allowed to pick up passengers in outer boroughs. \url{www1.nyc.gov/site/tlc/about/tlc-trip-record-data.page}} and thus comprises $p=4^2$ dimensional time-series sequences.
We used the original sequence for ten weeks and partitioned it into $y=\text{``weekday''}$ and $y=\text{``weekend''}$ episodes for each week, which resulted in a dataset comprising $n=20$ episodes (i.e., $n_\text{weekday}=n_\text{weekend}=10$).
As the data record the numbers of departures and arrivals within each 30 minutes, the length of each episode is $\tau_\text{weekday}=240$ or $\tau_\text{weekend} = 96$.
As preprocessing, we performed 24-point (i.e., half day) moving average and subtracted the mean value from each feature of each episode.

\subsubsection{Results}

We set $r=4$ and $\alpha=1$.
We examined the dynamic modes whose eigenvalues $\lambda$ roughly coincide with daily periodicity, which is one of the expected dynamical characteristics of the traffic.
In \cref{fig:taxi}, we exemplify such dynamic modes extracted by the discriminant DMD from a pair of weekday and weekend episodes.
We also illustrate the corresponding temporal profiles, $\vert \varphi_t \vert = \vert \lambda^{t-1} (\bm{z}^\ct \bm{x}_1) \vert$ (see \cref{eq:dmd}) in \cref{fig:taxi}.
In the weekend (\cref{fig:taxi:weekend}), almost every traffic except between Brooklyn and Bronx exhibits large amplitudes with the daily periodicity. In contrast, in the weekdays (\cref{fig:taxi:weekday}), the movements between Brooklyn and Queens and ones from them to Bronx have larger amplitudes.
Though directly interpreting such observations requires intensive domain knowledge, they can be useful, for example, for understanding the mechanism of the traffic and for making road-building plans.



\subsection{EEG Data}

\begin{figure}[t]
    \hspace{15mm}
    \subfloat[left-hand]{\label{fig:eeg:left}
        \def\arraystretch{0.6}
        \begin{tabular}[b]{l}
            \includegraphics[clip,height=4cm]{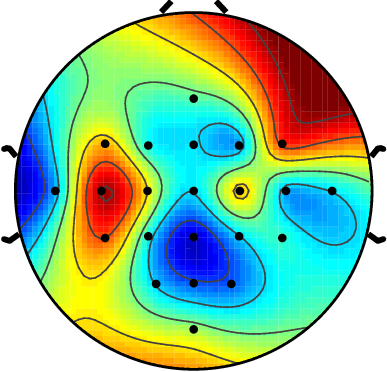} \\
            \pgfplotsset{width=3cm, height=0.45cm, compat=1.9}
            \begin{tikzpicture}
                \begin{axis}[
                    axis lines = left,
                    xtick = \empty, xticklabels = {,,},
                    ymin = -1, ymax = 1,
                    ytick = \empty, yticklabels = {,,},
                    ylabel = $\varphi$,
                    font = \tiny, enlargelimits = false, scale only axis]
                    \addplot[domain=0:10*2*pi, samples=200, thick]{-exp(-0.018*x)*sin(deg(x)+0.3*pi)};
                \end{axis}
            \end{tikzpicture} \\
            \vspace{0.2ex} \\
            \includegraphics[clip,height=4cm]{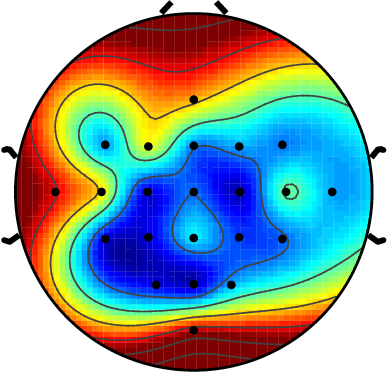} \\
            \pgfplotsset{width=3cm, height=0.45cm, compat=1.9}
            \begin{tikzpicture}
                \begin{axis}[
                    axis lines = left,
                    xtick = \empty, xticklabels = {,,},
                    ymin = -0.2, ymax = 3.7,
                    ytick = \empty, yticklabels = {,,},
                    ylabel = $\varphi$,
                    font = \tiny, enlargelimits = false, scale only axis]
                    \addplot[domain=0:1, samples=200, thick]{3.7*exp(-20*x)};
                \end{axis}
            \end{tikzpicture}
        \end{tabular}
    }
    \hspace{8mm}
    \subfloat[right-hand]{\label{fig:eeg:right}
        \def\arraystretch{0.6}
        \begin{tabular}[b]{l}
            \includegraphics[clip,height=4cm]{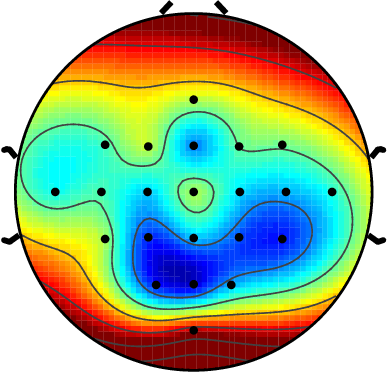} \\
            \pgfplotsset{width=3cm, height=0.45cm, compat=1.9}
            \begin{tikzpicture}
                \begin{axis}[
                    axis lines = left,
                    xtick = \empty, xticklabels = {,,},
                    ymin = -1, ymax = 1,
                    ytick = \empty, yticklabels = {,,},
                    ylabel = $\varphi$,
                    font = \tiny, enlargelimits = false, scale only axis]
                    \addplot[domain=0:4.8*2*pi, samples=200, thick]{exp(-0.016*x)*cos(deg(x)+0.02*pi)};
                \end{axis}
            \end{tikzpicture} \\
            \vspace{0.2ex} \\
            \includegraphics[clip,height=4cm]{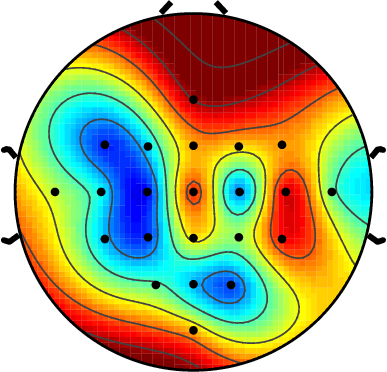} \\
            \pgfplotsset{width=3cm, height=0.45cm, compat=1.9}
            \begin{tikzpicture}
                \begin{axis}[
                    axis lines = left,
                    xtick = \empty, xticklabels = {,,},
                    ymin = -6.2, ymax = 0.2,
                    ytick = \empty, yticklabels = {,,},
                    ylabel = $\varphi$,
                    font = \tiny, enlargelimits = false, scale only axis]
                    \addplot[domain=0:1, samples=200, thick]{-6.2*exp(-20*x)};
                    \node[inner sep=0pt, anchor=south east] at (axis cs: 1,-5.5) {\tiny 1 [sec]};
                \end{axis}
            \end{tikzpicture}
        \end{tabular}
    }
    \hspace{2mm}
    \begin{tikzpicture}
        \node[inner sep=0pt, anchor=south east] at (0,0) {\includegraphics[clip,height=4.4cm]{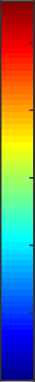}};
        \node[inner sep=0pt, anchor=south west] at (0.1, 0) {\scriptsize 0};
        \node[inner sep=0pt, anchor=north west] at (0.1, 4.4) {\scriptsize +};
        \node[inner sep=0pt, rotate=90] at (0.2,2.2) {\tiny magnitude of dynamic modes};
    \end{tikzpicture}
    \caption{Examples of the distinctive coherent patterns and their dynamics extracted from the EEG data. Representative ones are shown for each of the two labels: (a) left- and (b) right-hand motor imagery.
    The patterns in the upper row similarly follow the slightly decaying oscillations, and ones in the lower row follow the rapidly converging dynamics.
    }
    \label{fig:eeg}
\end{figure}
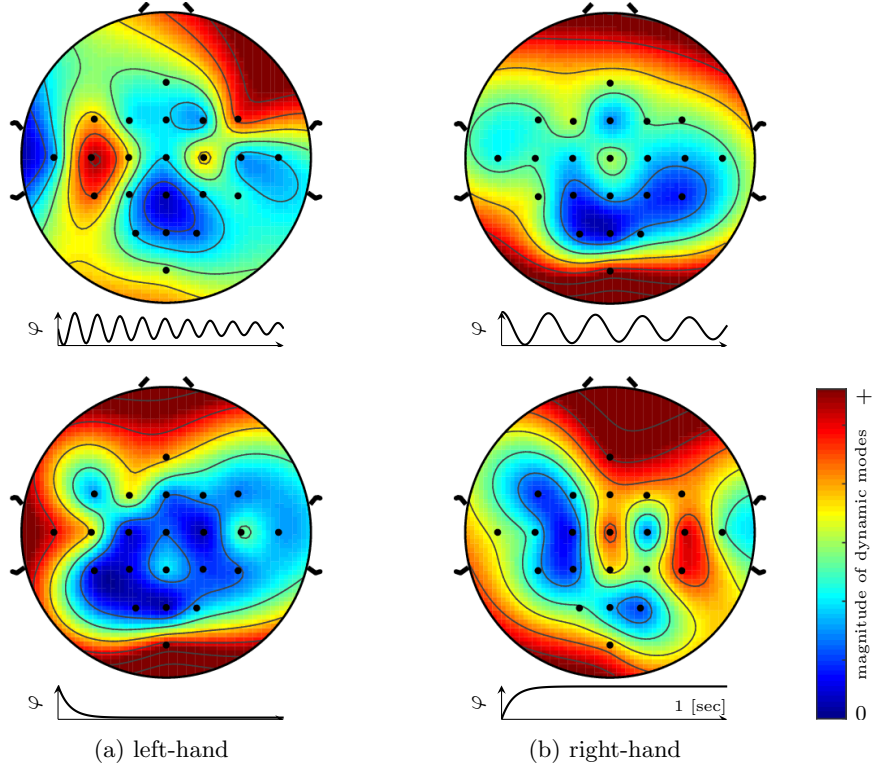

\subsubsection{Dataset}

We applied the discriminant DMD to electroencephalography (EEG) data to show a possibility of application in such a biological domain.
We used the dataset publicly provided in BCI Competition IV 2a\footnote{\url{www.bbci.de/competition/iv/}}.
The original data was recorded using twenty-two Ag/AgCl electrodes with sampling rate of 250 [Hz].
We eliminated noises by applying the band-pass filter between 0.5 [Hz] and 100 [Hz] and the spatial filter with the surface Laplacian \citep{mcfarland1997spatial}.
We used ten episodes of the same subject behaving two motor imagery tasks: imagination of movement of the left hand or the right hand.
Hence, the dataset comprises $n=20$ episodes (i.e., $n_\text{left-hand}=n_\text{right-hand}=10$).
For each sequence, we used only a part of length $\tau=250$ (corresponding to $1$ [sec]) from one second after the task started.

\subsubsection{Results}

We set $r=4$ and $\alpha=1$.
As with the previous experiment, we show representative examples of distinctive dynamic modes and their temporal profiles in \cref{fig:eeg}.
While such information alone is not necessarily sufficient for analyzing EEG signals, it may be informative when combined with existing domain knowledge.
For example, we may find some similarity between the patterns in \cref{fig:eeg} and the ones obtained in the previous work \citep{Blankertz08} using the method of common spatial pattern.

Another interesting observation here is that those pairs of distinctive patterns have different dynamical properties.
That is, in \cref{fig:eeg}, the upper-row patterns similarly follow slightly decaying oscillations, and the lower-row patterns have the rapidly converging temporal profiles.
One of the advantages of the proposed method is that it can extract dynamics of spatial coherent patterns as shown in this example.


\section{Discussion}
\label{discussion}

The numerical results in the previous section show the utility of the proposed method, discriminant DMD, for extracting sets of distinctive coherent patterns (i.e., oscillating spatial patterns) that well reconstruct time-series signals \emph{and} distinguish labels.
Such information extraction is useful in applications where we want to understand data characteristics, craft features for classification and regression, and design controllers.
For example, the distinctive patterns of house temperature (\cref{expt:house}) can be utilized in designing the location and controllers of air conditioners.
Analysis of human motion capture data with label information (\cref{expt:mocap}) can help more precise data-driven understanding of human locomotion (see, e.g., \citep{Fujii19locom}).
Nonetheless, the discriminant DMD itself can hardly be sufficient for \emph{solving} some task completely, as so are any other data analysis methods such as PCA and DMD.
We need to combine the discriminant DMD with other subsequent methodologies for more practical utility, and investigating such possibility in various domains is a promising direction of future study.

One of notable limitations of the current discriminant DMD is that it does not provide a principled way to determine which dynamic modes are \emph{the most} important both for representing sequences and distinguishing labels.
As we define the similarity between episodes via a kernel function on \emph{sets} of dynamic modes, it is usually challenging to tell exactly which pairs of dynamic modes are solely to be focused out of $r$ modes computed by the method.
Hence, one of the primal usages of the current method will be in exploratory data analysis, where we can interpret the results based on our domain knowledge.
For example, in \cref{expt:taxi}, we picked up the pair of dynamic modes based on the knowledge that the data should indicate daily periodicity as its characteristics.
However, it will be more useful if there is a principled way to rank importance of dynamic modes for data analysis with less prior knowledge.


\section{Conclusion}
\label{concl}

In this work, we developed a method for discovering spatio-temporal coherent patterns from labeled data collections, namely \emph{discriminant DMD}.
The discriminant DMD computes distinctive coherent patterns that contribute to major differences of dynamics with different labels by optimizing the objective that takes the reconstruction goodness of DMD and the class-separation goodness of discriminant analysis into account.
We have demonstrated applications of the discriminant DMD using four different types of real-world datasets, with which we empirically validated that the proposed method extracts spatial patterns that well reconstruct data \emph{and} distinguish different labels.
Such pattern extraction is useful for exploratory data analysis towards understanding spatio-temporal data.
Important directions of future research include the investigation of the utility of the discriminant DMD in further subsequent tasks such as feature crafting and controller design.


\appendix

\section{Gradient of Objective Function}
\label{appendix:grad}

The gradient of the objective function in \eqref{eq:problem}, $f$, with regard to $\theta_{i,j}$ (for $i=1,\dots,n$ and $j=1,\dots,r$), is obtained as follows.
The gradient is computed from derivatives of $f$ as
\begin{equation}
    \nabla_{\theta_{i,j}} f = 2 \frac{\partial f}{\partial \bar\theta_{i,j}} = 2 \overline{\frac{\partial f}{\partial \theta_{i,j}}},
\end{equation}
where $\bar{\cdot}$ means the complex conjugate.
The derivative is
\begin{equation}\label{eq:deriv1}
    \frac{\partial f}{\partial \theta_{i,j}}
    =
    \frac{\partial}{\partial \theta_{i,j}} \left( \frac{f_\text{DMD}(\Theta_{1:n})}{f_\text{KFD}(\Theta_{1:n})^\alpha + \epsilon} \right)
    =
    \frac{
        (f_\text{KFD}^\alpha + \epsilon) \frac{\partial f_\text{DMD}}{\partial \theta_{i,j}}
        - \alpha f_\text{DMD} f_\text{KFD}^{\alpha-1} \frac{\partial f_\text{KFD}}{\partial \theta_{i,j}}
    }{\big( f_\text{KFD}^\alpha + \epsilon \big)^2}.
\end{equation}

The first of the two derivatives in \eqref{eq:deriv1}, $\frac{\partial f_\text{DMD}}{\partial \theta_{i,j}}$ is
\begin{equation}\label{eq:deriv2}
    \frac{\partial f_\text{DMD}}{\partial \theta_{i,j}}
    = \operatorname{sum}
    \left(
        \frac{\partial f_\text{DMD}}{\partial \bm{V}_{\Theta_i}} \circ
        \frac{\partial \bm{V}_{\Theta_i}}{\partial \theta_{i,j}}
    \right),
\end{equation}
where $\operatorname{sum}(\bm{A})$ denotes the summation of all the elements of matrix $\bm{A}$, and
\begin{equation}
    \frac{\partial f_\text{DMD}}{\partial \bm{V}_{\Theta_i}}
    =
    \frac1{n \tau_i}
    \Big(
        (\bm{V}_{\Theta_i}^\pinv \bm{V}_{\Theta_i} - \bm{I}) \bm{X}_i^\ct\bm{X}_i \bm{V}_{\Theta_i}^\pinv
    \Big)^\tr,
\end{equation}
and
\begin{equation}\label{eq:deriv_V}
    \frac{\partial \bm{V}_{\Theta_i}}{\partial \theta_{i,j}}
    =
    \begin{bmatrix}
        0 & 0 & 0 & \cdots & 0 \\
        \vdots & \vdots & \vdots & & \vdots \\
        0 & 1 & 2\theta_j & \cdots & (\tau_i-1)\theta_j^{\tau_i-2} \\
        \vdots & \vdots & \vdots & & \vdots \\
        0 & 0 & 0 & \cdots & 0
    \end{bmatrix}.
\end{equation}

The other derivative in \eqref{eq:deriv1}, $\frac{\partial f_\text{KFD}}{\partial \theta_{i,j}}$, is obviously
\begin{equation}\label{eq:deriv3}
    \frac{\partial f_\text{KFD}}{\partial \theta_{i,j}}
    =
    \frac{\partial Q_1}{\partial \theta_{i,j}} Q_2
    + \frac{\partial Q_2}{\partial \theta_{i,j}} Q_1.
\end{equation}
Below we basically follow the deviation by You \emph{et al.} \citep{You11} to complete this calculation.

As for $Q_1$ in \eqref{eq:deriv3}, we have
\begin{multline}
    \frac{\partial Q_1}{\partial \theta_{i,j}} = \frac2{c(c-1)} \sum_{l=1}^{c-1} \sum_{m=l+1}^c
    \left\{
        \frac{
            \frac{\partial \trace(\bm\Sigma_l^\phi \bm\Sigma_m^\phi)}{\partial \theta_{i,j}} \Big( \trace(\bm\Sigma_l^\phi \bm\Sigma_l^\phi) + \trace(\bm\Sigma_m^\phi \bm\Sigma_m^\phi) \Big)
        }{
            \Big( \trace(\bm\Sigma_l^\phi \bm\Sigma_l^\phi) + \trace(\bm\Sigma_m^\phi \bm\Sigma_m^\phi) \Big)^2
        }
    \right.
        \\
    \left.
        -
        \frac{
            \trace(\bm\Sigma_l^\phi \bm\Sigma_m^\phi) \Big( \frac{\partial \trace(\bm\Sigma_l^\phi \bm\Sigma_l^\phi)}{\partial \theta_{i,j}} + \frac{\partial \trace(\bm\Sigma_m^\phi \bm\Sigma_m^\phi)}{\partial \theta_{i,j}} \Big)
        }{
            \Big( \trace(\bm\Sigma_l^\phi \bm\Sigma_l^\phi) + \trace(\bm\Sigma_m^\phi \bm\Sigma_m^\phi) \Big)^2
        }
    \right\},
\end{multline}
where $\bm\Sigma_l^\phi$ denotes the sample covariance matrix of the $l$-th class ($l \in [1,c]$) in the feature space as in section~\ref{back:kfd}.

Recall that we gave an index to the sequences in a dataset $\mathcal{D}$ from $1$ to $n$ (see \eqref{eq:dataset}), which was indexed by subscript $i$.
Suppose a subset of $\mathcal{D}$ comprising sequences with label $l$, that is,
\begin{equation}
    \mathcal{D}_l = \{ (X_i, y_i) \mid i=1,\dots,n, \, y_i = l \},
\end{equation}
and let $n_l=\vert\mathcal{D}_l\vert$ denote the number of sequences with label $l$.
Now suppose to give an arbitrary order to the elements of $\mathcal{D}_l$.
We denote the \emph{original index} (in $\mathcal{D}$) of the $p$-th element of $\mathcal{D}_l$ by $i_{l,p}$.
In other words, $i_{l,p}$ denotes the index of the $p$-th (in an arbitrary order) sequence with label $l$.
Given such definitions, we denote the value of the kernel function $k_\text{DMS}$ in \eqref{eq:kdms} between $(\Theta_{i_{l,p}}, \bm{X}_{i_{l,p}})$ and $(\Theta_{i_{m,q}}, \bm{X}_{i_{m,q}})$ by $K_{l,m}^{p,q}$, that is,
\begin{equation}
    K_{l,m}^{p,q} = k_{\text{DMS}}(\mathcal{W}_{\Theta_{i_{l,p}}, \bm{X}_{i_{l,p}}}, \mathcal{W}_{\Theta_{i_{m,q}}, \bm{X}_{i_{m,q}}}).
\end{equation}
Moreover, we let $\bm{K}_{l,m}$ be the kernel matrix whose $(p,q)$-element is $K_{l,m}^{p,q}$.

Given the above quantities, we can compute the quantities related to $\bm\Sigma^\phi$ as follows:
\begin{equation}
    \trace(\bm\Sigma_l^\phi \bm\Sigma_m^\phi)
    = \trace (
        \underbrace{\bm{K}_{l,m}(\bm{I}-\bm{O}_{n_m})}_{\tilde{\bm{K}}_{l,m}}
        \underbrace{\bm{K}_{m,l}(\bm{I}-\bm{O}_{n_l})}_{\tilde{\bm{K}}_{m,l}}
    )
    = \sum_{p=1}^{n_l} \sum_{q=1}^{n_m} \tilde{K}_{l,m}^{p,q} \tilde{K}_{m,l}^{q,p},
\end{equation}
and
\begin{multline}
    \frac{\partial \trace(\bm\Sigma_l^\phi \bm\Sigma_m^\phi)}{\partial \theta_{i,j}}
    = \sum_{p} \sum_{q}
    \Bigg\{
        \Bigg(
            \frac{\partial K_{l,m}^{p,q}}{\partial \theta_{i,j}} - \frac1{n_m}\sum_{q} \frac{\partial K_{l,m}^{p,q}}{\partial \theta_{i,j}}
        \Bigg) \tilde{K}_{m,l}^{q,p}
        \\
        \qquad\qquad
        + \tilde{K}_{l,m}^{p,q} \Bigg(
            \frac{\partial K_{m,l}^{q,p}}{\partial \theta_{i,j}} - \frac1{n_l}\sum_{p} \frac{\partial K_{m,l}^{q,p}}{\partial \theta_{i,j}}
        \Bigg)
    \Bigg\},
\end{multline}
where $\bm{O}_{n_l}$ denotes an $n_l \times n_l$ matrix with all elements being $1/n_l$.

The remaining part of \eqref{eq:deriv3} is computed by
\begin{multline}
    \frac{\partial Q_2}{\partial \theta_{i,j}}
    = \sum_{l=1}^{c-1} \sum_{m=l+1}^c \frac{n_l n_m}{n^2}
    \Bigg(
        \frac1{n_l^2} \sum_q \sum_{q'} \frac{\partial K_{l,l}^{q,q'}}{\partial \theta_{i,j}}
        \\
        - \frac2{n_l n_m} \sum_q \sum_p \frac{\partial K_{m,l}^{q,p}}{\partial \theta_{i,j}}
        + \frac1{n_m^2} \sum_p \sum_{p'} \frac{\partial K_{m,m}^{p,p'}}{\partial \theta_{i,j}}
    \Bigg).
\end{multline}

The final missing piece is the derivative of kernel, $\frac{\partial K_{l,m}^{p,q}}{\theta_{i,j}}$, which is computed as
\begin{equation}
    \frac{\partial K_{i,k}^{p,q}}{\partial \theta_{i,j}}
    = \operatorname{sum}
    \left(
        \frac{\partial K_{i,k}^{p,q}}{\partial \bm{V}_s}
        \circ \frac{\partial \bm{V}_s}{\partial \theta_{i,j}}
    \right).
\end{equation}
As the second term is already given in \eqref{eq:deriv_V}, we focus on the first term.
Let $\bm{A}_i \in \mathbb{C}^{p \times r}$ and $\bm{B}_i \in \mathbb{C}^{p \times r}$ denote the matrices whose columns comprise left- and right-singular vectors of $\bm{X}_i \bm{V}_{\Theta_i}^\pinv$.
Let $\bm{S}_i$ be a diagonal matrix comprising the singular values of $\bm{X}_i \bm{V}_{\Theta_i}^\pinv$ (in the same order with $\bm{A}_i$ and $\bm{B}_i$).
Then,
\begin{multline}
    \frac{\partial K_{l,m}^{p,q}}{\partial \bm{V}_{\Theta_i}}
    = 
    \Big(
        \bm{V}_{\Theta_i}^\pinv (\bm{V}_{\Theta_i}^\pinv)^\ct (\bm{C}_{l,m}^{p,q})^\ct (\bm{I}-\bm{V}_{\Theta_i}\bm{V}_{\Theta_i}^\pinv)^\ct
        \\
        + (\bm{I}-\bm{V}_{\Theta_i}^\pinv\bm{V}_{\Theta_i})^\ct (\bm{C}_{l,m}^{p,q})^\ct (\bm{V}_{\Theta_i}^\pinv)^\ct \bm{V}_{\Theta_i}^\pinv
        - \bm{V}_{\Theta_i}^\pinv \bm{C}_{l,m}^{p,q} \bm{V}_{\Theta_i}^\pinv
    \Big)^\tr,
\end{multline}
where
\begin{equation}
    \bm{C}_{l,m}^{p,q} = (\bm{B}_{i_{l,p}} \bm{S}_{i_{l,p}}^{-1} \bm{A}_{i_{l,p}}^\ct) \bm{A}_{i_{m,q}}\bm{A}_{i_{m,q}}^\ct (\bm{I}-\bm{A}_{i_{l,p}}\bm{A}_{i_{l,p}}^\ct) \bm{Y}_{i_{l,p}}.
\end{equation}


\bibliographystyle{siamplain}
\bibliography{main}

\end{document}